\documentclass[sn-mathphys]{sn-jnl}

\jyear{2021}%

\theoremstyle{thmstyleone}%
\theoremstyle{thmstyletwo}%
\usepackage{epstopdf}

\usepackage{graphicx}	
\usepackage{float}
\usepackage{color}

\usepackage{booktabs}
\usepackage{threeparttable}
\usepackage[section]{placeins}
\usepackage{caption}
\usepackage{tabularx}
\usepackage{booktabs}
\usepackage{amsmath,amsfonts,amssymb,bm}
\usepackage{multirow}
\usepackage{diagbox}
\usepackage{natbib}

\usepackage{subfigure} 				
\usepackage{autobreak} 
\theoremstyle{thmstylethree}%
\begin{document}

\title[Article Title]{Heat Source Layout Optimization Using Automatic Deep Learning Surrogate and Multimodal Neighborhood Search Algorithm}


\author[1]{\fnm{Jialiang} \sur{Sun}}\email{sun1903676706@163.com}

\author[2]{\fnm{Xiaohu} \sur{Zheng}}\email{zhengboy320@163.com}

\author*[1]{\fnm{Wen} \sur{Yao}}\email{wendy0782@126.com}

\author[1]{\fnm{Xiaoya} \sur{Zhang}}\email{zhangxiaoya09@nudt.edu.cn}

\author*[1]{\fnm{Weien} \sur{Zhou}}\email{weienzhou@outlook.com}

\author[1]{\fnm{Xiaoqian} \sur{Chen}}\email{chenxiaoqian@nudt.edu.cn}

\affil*[1]{\orgdiv{Defense Innovation Institute}, \orgname{Chinese Academy of Military Science}, \city{Beijing}, \postcode{100000},  \country{China}}

\affil[2]{\orgdiv{College of Aerospace Science and Engineering}, \orgname{National University of Defense Technology}, \orgaddress{\city{Changsha}, \postcode{410073}, \country{China}}}


\abstract{Deep learning surrogate assisted heat source layout optimization (HSLO) could learn the mapping from layout to its corresponding temperature field, so as to substitute the simulation during optimization to decrease the computational cost largely. However, it faces two main challenges: 1) the neural network surrogate for the certain task is often manually designed to be complex and requires rich debugging experience, which is challenging for the designers in the engineering field; 2) existing algorithms for HSLO could only obtain a near optimal solution in single optimization and are easily trapped in local optimum. To address the first challenge, considering reducing the total parameter numbers and ensuring the similar accuracy as well as, a neural architecture search (NAS) method combined with Feature Pyramid Network (FPN) framework is developed to realize the purpose of automatically searching for a small deep learning surrogate model for HSLO. To address the second challenge, a multimodal neighborhood search based layout optimization algorithm (MNSLO) is proposed, which could obtain more and better approximate optimal design schemes simultaneously in single optimization. Finally, two typical two-dimensional heat conduction optimization problems are utilized to demonstrate the effectiveness of the proposed method. With the  similar  accuracy, NAS finds models with 80$\%$ fewer parameters, 64$\%$ fewer FLOPs and 36$\%$ faster inference time than the original FPN. Besides, with the assistance of deep learning surrogate by automatic search, MNSLO could achieve multiple near optimal design schemes simultaneously to provide more design diversities for designers. }




\maketitle

\section{Introduction}\label{sec1}

Layout design \cite{Chen2018_a} is the key step in the process of whole satellite design \cite{Zheng2019Complex,Zheng2019,Zheng2020}. In practical engineering, as the heat intensity and size  of the electronics components become higher and smaller, satellite heat source layout optimiztaion (HSLO) has been a challenging concern \cite{Chen2020}, which intends to decrease the maximum temperature in the layout domain \cite{Chen,Chen2,Chen3,Aslan}.

 HSLO needs to take substantial heat simulation during optimization, which brings great calculation burden. To realize the acceleration, deep learning surrogate assisted method was first developed by Chen et al \cite{Chen2020}. They utilized the Feature Pyramid Network (FPN) as a predictor to learn the mapping from the heat source layout to the corresponding temperature field. Further, they developed a neighborhood optimization algorithm to obtain the near optimal layout design. Although their method proves to be feasible, there exist two main challenges:

\textbf{1) The deep learning surrogate is usually manually designed to be complex and requires rich engineering experience, which is challenging for the designers.}

\textbf{2) Due to the multimodality and high dimensionality of the HSLO problem, the algorithm is easily trapped in local optimum.}

As for the first challenge, neural architecture search (NAS) has made great success in computer vision tasks, which could automatically search for the architecture of neural network with high performance \cite{Zoph2018,Liu2018,Zhong2018}. The main procedure of NAS includes three steps. The first step is to define a suitable search space. Then the evaluation metric to assess the performance of candidate model architecture is needed to be determined. At last, the efficient searche strategy is utilized to find a better model architecture. Early work about NAS searched the optimal model architectures using reinforcing learning or pure evolutionary algorithms, which brings unbearable computational burden \cite{Zoph2018}. To realize the purpose of acceleration, one shot neural architecture search \cite{Pham2018,chu2020mixpath} and differential architecture search (DARTS) \cite{Liudarts2018} turn to be popular. The former mainly consists of two steps, including training a supernet including all operations and evolutionary search for the optimal path based on the supernet. The latter greatly enhanced the efficiency by gradient-based search \cite{jin2019rcdarts,xu2019pcdarts,chen2019progressive}. Currently, the work about NAS for FPN model could be seen in \cite{HangAuto,Golnaz2019}, which tried to find the near optimal FPN model architectures by evolutionary algorithms or gradient-based methods in object detection. However, both of them are not directly suitable for the temperature field prediction task. First, the search process of them need to take over 30 GPU-days. Second, it shows that the loss value of DARTS could not decrease in HSLO task in our experiments. To address these challenges, we utilize the multi-objective neural architecture search to adaptively construct the deep learning surrogate at low cost, which can learn the mapping from the layout to the temperature field well. The main process consists of two steps: training a supernet including all possible sub-model architectures and searching the near optimal architectures by the nondominated sorting genetic algorithm (NSGA-II) \cite{Seshadrinsga}.


As for the second challenge, multimodal optimization,  which seeks multiple  optima simultaneously, has attracted much attention in recent years \cite{qu2020self}. Many practical engineering problems are multimodal such as electromagnetic design \cite{Ling2005}, data mining \cite{sheng1997,Li2009, li2022approximated} and layout design \cite{Chen2018_a}. So the researchers hope to obtain as many optimal solutions as possible in the global optimization. Most of existing algorithms are based on clustering or niching strategy \cite{Haghbayan2017, Rim2016}, which could partition the whole population into multiple groups. Then each group is responsible for detecting the promising area of design space. Eventually, the population would converge to multiple solutions, reaching the purpose of multimodal optimization. However, most of the previous work about multimodal optimization focus on continuous problems. So it is general to use euclid distance to divide the whole population. Although few work of discrete multimodal optimiation could be seen in \cite{huang2019tevc} to solve traveling salesman problem, the algorithm could not be directly used in the discrete HSLO problem. Thus it is necessary to design the suitable strategy to conduct multimodal optimization according to the characteristic of discrete HSLO problem. To realize it, we first define a similarity metric to evaluate the distance of diferent layout schemes in the discrete domain. Then, we could cluster the population into multiple groups to preserve diversity. At last, taking the best individual of each group as the initial layout scheme, the neighborhood search strategy as a kind of local search is adopted to seek the optimal layout. In the process of local search, unlike only selecting the best layout into next iteration in previous work \cite{Chen2020}, the searched multiple optimal solutions are preserved to an archieve in every iteration, so as to output multiple solutions finally. 

The preliminary version of this paper appeared as \cite{9368601}, where NAS is employed to construct the surrogate in one simple HSLO case. In this paper, we propose the framework of multimodal heat source layout optimization design based on neural architecture search (MHSLO-NAS). The contributions could be concluded as follows:

\begin{itemize}
	\item We develop a multi-objective neural architecture search method for heat source layout optimization, which could realize the purpose of automatically searching for a better deep learning surrogate for learning the mapping from layout to temperature field.
	
	\item{The searched model architecture by NAS yields the state-of-art performance compared with the previous hand-crafted. With the  similar arruracy, we can find models with 80$\%$ fewer parameters, 64$\%$ fewer FLOPs and 36$\%$ faster inference time than the original FPN. }
	
	\item{We propose a multimodal neighborhood search based layout optimization algorithm to solve HSLO based on the searched model, which could obtain multiple near optimal solutions simultaneously to provide more design choices for designers. And we achieve the state-of-art optimal layout schemes on both of two cases compared with other algorithms. }
	%
\end{itemize}

The remainder of this paper is organized as follows. In Section \ref{sec2}, the mathematical model of HSLO design problem is constructed and the deep learning assisted HSLO method is demonstrated briefly. Then in Section \ref{sec3}, the proposed MHSLO-NAS framework by us is elaborated. In Section \ref{sec5}, the multi-objective neural architecture search for HSLO is introduced from the definition of search space, search strategy and performance evaluation in detail. In Section \ref{sec6}, a novel multimodal neighborhood search based layout optimization algorithm is introduced. In Section\ref{sec7}, the effectiveness of our proposed method is verified on two cases. The solutions solved by MHSLO-NAS is evaluated from two aspects: the performance of the model searched  by NAS and the optimal layout schemes obtained by MNSLO. Finally, the conclusions are discussed in Section \ref{sec8}.

\section{Results}\label{sec2}
\subsection{ Problem description}

\begin{figure}[htb]
	\centering
	\subfigure[VP problem]{
		\includegraphics[height=0.32\linewidth]{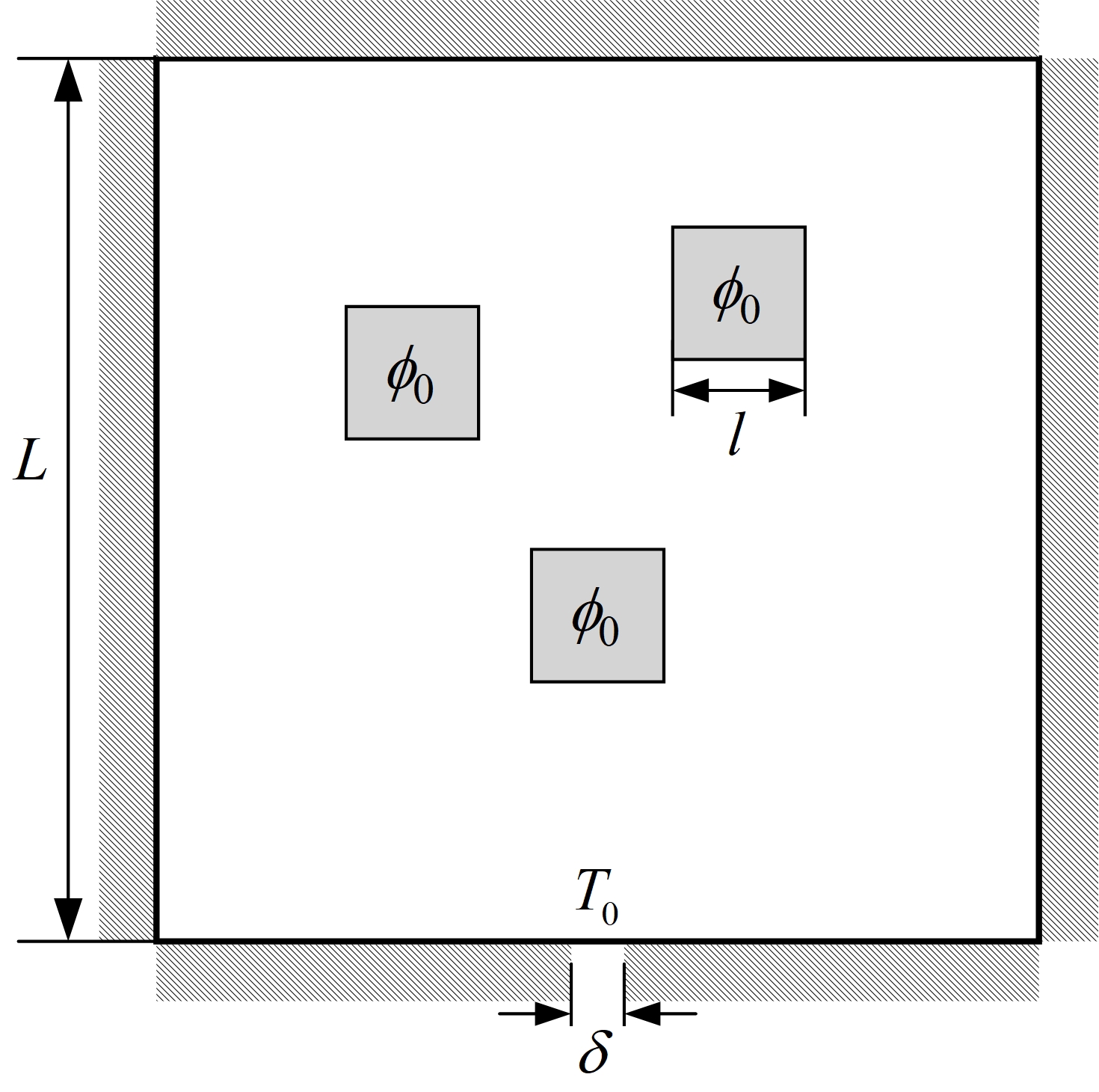}
	}
	\subfigure[Data preparation]{
		\includegraphics[height=0.32\linewidth]{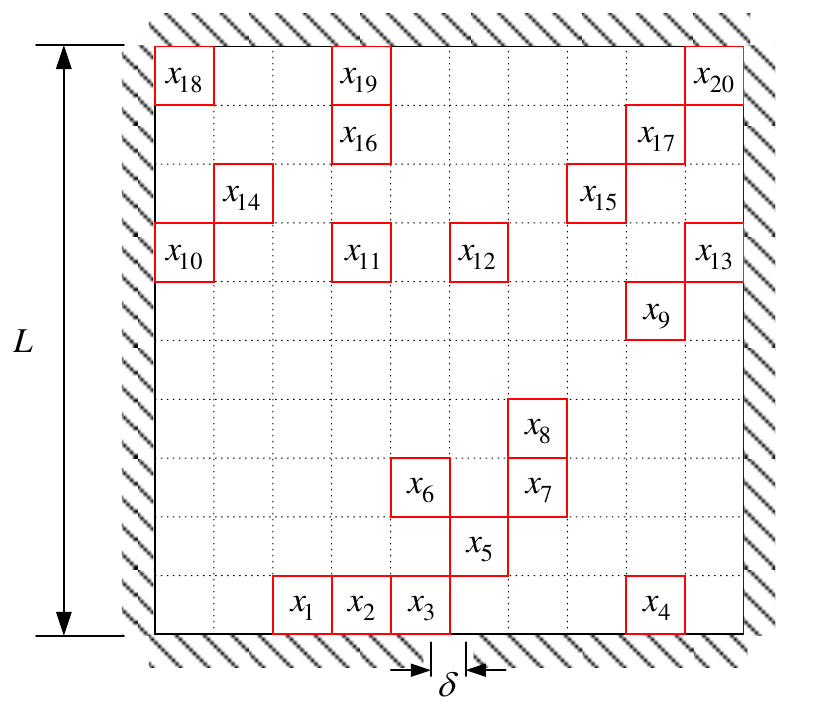}
	}
	\label{datapreparation}
	\caption{Problem definition \cite{Chen2020}.}
	\label{vp}
\end{figure}

HSLO aims to obtain the near optimal layout design that minimizes the maximum temperature in specific layout domain. In previous work \cite{Chen,Chen2,Chen3,Aslan,Chen2020}, the volume-to-point (VP) heat conduction problem is taken as an example, which is presented in Figure~\ref{vp}(a). Multiple heat sources are placed in a square domain, where all the boundaries except one tiny heat sink ($ T_0 $)  are adiabatic.

The temperature field ($T$) generated by heat sources in the layout domain can be obtained by solving the Poisson's equation as follows:
\begin{equation}
\begin{split}
& \frac{\partial}{\partial x}\left(k \frac{\partial T}{\partial x}\right)+\frac{\partial}{\partial y}\left(k \frac{\partial T}{\partial y}\right)+\phi(x, y)=0 \\
& { \quad T=T_{0} \quad \text{,} \quad 
	k \frac{\partial T}{\partial \mathbf{n}}=0 \quad }
\label{eq1}
\end{split}
\end{equation}
where $\phi(x,y)$ denotes the intensity distribution function of heat sources.

The positions of heat sources could determine $\phi(x,y)$, which can be described as
\begin{equation}
\phi (x,y)=\left\{ 
\begin{aligned}
& {{\phi }_{0}}, & & (x,y)\in \Gamma   \\ 
& 0, 			 & & (x,y)\notin \Gamma   \\
\end{aligned}
\right.
\label{eq2}
\end{equation}
where $\phi_0$ denotes the intensity of single heat source and $\Gamma$ stands for the area where the heat source is placed.

The thermal performance of the heat source layout design is assessed using the maximum temperature  ($T_{max}$) in the layout domain. The normalized version of the metric can be described as 
\begin{equation}
R_m = \frac{T_{max} - T_0}{\phi_{0} L^2 /k}
\end{equation}


Considering the basic non-overlapping constraint:
\begin{equation}
\begin{aligned}
& \Gamma_i \cap \Gamma_j = \emptyset & & \forall i\ne j \\
& \Gamma_i \subset \Gamma_0  & & \forall i = 1,2,...,N_s
\end{aligned}
\end{equation}
where $\Gamma_i$ denotes the area of the $i$th heat source that is placed and $\Gamma_0$ stands for the whole layout domain. $N_s$ denotes the number of heat sources that are palced.

To sum up, the mathematical model for the HSLO problem in this paper can be described as
\begin{equation}
\left \{
\begin{aligned}
& \text{find} & & \bm{X} \\
& \text{minimize}    & & R_m \\
& s.t. 	& & \Gamma_i \cap \Gamma_j = \emptyset & & \forall i\ne j \\
& 		& & \Gamma_i \subset \Gamma_0  & & \forall i = 1,2,...,N_s
\end{aligned}
\right.
\label{eq5}
\end{equation}
where $\bm{X}$ represents the heat source layout scheme. 

\subsection{HSLO using deep learning }\label{sec2b}

\begin{figure}[h]
	\centering
	\includegraphics[width=0.55\linewidth]{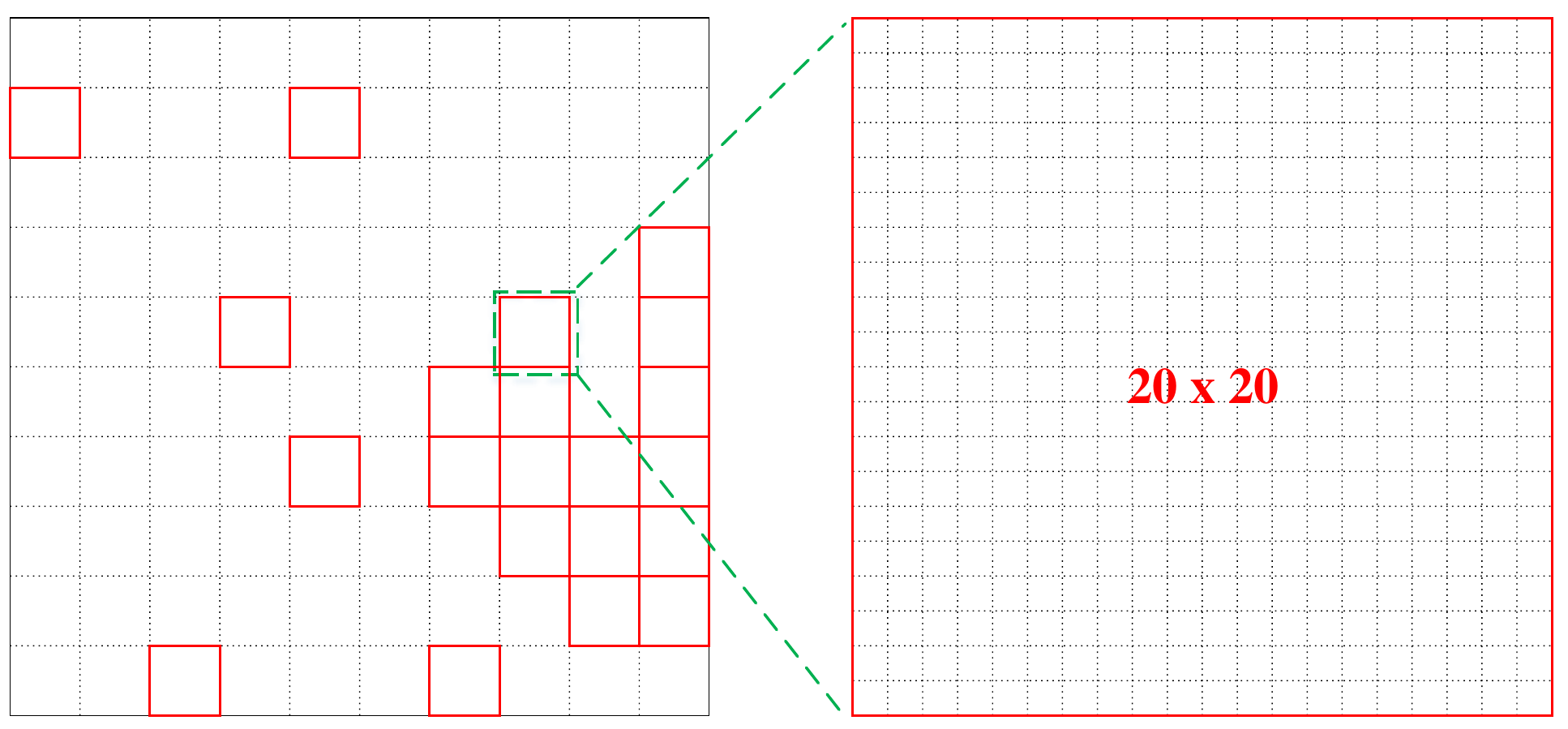}
	\caption{The illustration of layout representation \cite{Chen2020}.}\label{grid}
\end{figure}

In previous work \cite{Chen2020}, the defined problem is a discrete layout design problem. In detail, as illustrated in Figure \ref{vp}(b), 20 heat sources that share the same size with the length of $l = 0.01 \text{m}$ and intensity of $\phi_0=10000$W/m$^2$ are placed in a $200\times 200$ square domain with the length of $L=0.1$m, which is divided uniformly into $10\times 10$, resulting in 100 cells. Each heat source could only be placed in the discrete $20 \times 20$ plane, which is illustrated in Figure \ref{grid}. The thermal conductivity is set as $k=$1 W/(m$ \cdot $K). The width and temperature value of the narrow heat sink are set as $\delta=0.001$m and constant at $T_0 = 298$K respectively. Then these settings are combined into Eq.(\ref{eq5}) to form as a heat source layout optimization problem.

\begin{figure}[h]
	\centering
	\includegraphics[scale=0.35]{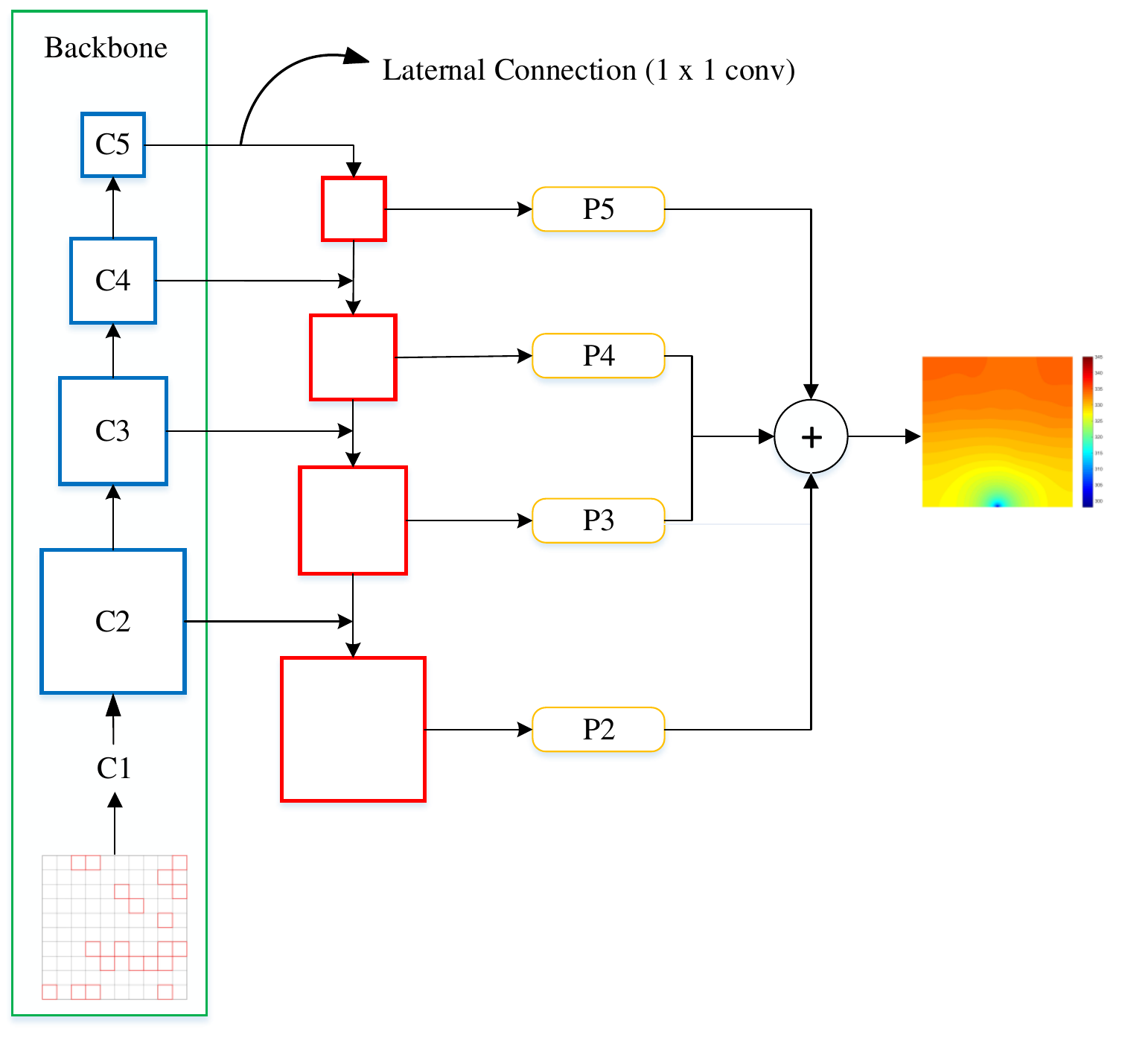}
	\caption{The overview of the architecture of FPN model \cite{9368601}.}
	\label{fpn}
\end{figure}

To solve above issue, Chen et al. \cite{Chen2020} first proposed to utilize FPN model to learn the mapping between layout and temperature field. After generating various layout schemes, they obtain the corresponding temperature fields that are numerically calculated by the finite-difference method (FDM) \cite{ReimerA2013}. The whole architecture of FPN model includes three parts, which is presented in Figure \ref{fpn}. The first part is the backbone network to extract the feature of input images. Given an input layout image with 200$\times$200 resolution, the backbone wolud reduce the image size halfly for six times after convolutional and maxpool operation.  As the left part of Figure \ref{fpn} shows, the resolutions of C1, C2, C3, C4 and C5 are 50 $\times$ 50, 25 $\times$ 25, 13 $\times$ 13, 7 $\times$ 7, 4 $\times$ 4 respectively. ResNet50 is utilized as the backbone in the original FPN model. The second part of FPN is top-down pathway. The low resolution image would be restored back to double. Thus the red box stands for feature maps with the size of 7 $\times$ 7, 13 $\times$ 13, 25 $\times$ 25, 50 $\times$ 50 respectively. In addition, the feature map in each red box woud be upsampled to 50$\times$50 separately, which output the feature map P1, P2, P3, P4.  The laternal connection is the third part to connect the neighboring feature maps in backbone and top-down pathway together. The final merged feature map would be upsampled to 200 $\times$ 200 resolution. Eventually, denote the final temperature field ouput as $T$, $T = P2 + P3 + P4 + P5$.

\section{The framework of MHSLO-NAS}\label{sec3}

%
%
%

\begin{figure*}[htbp]
	\centering
	\includegraphics[scale=0.5]{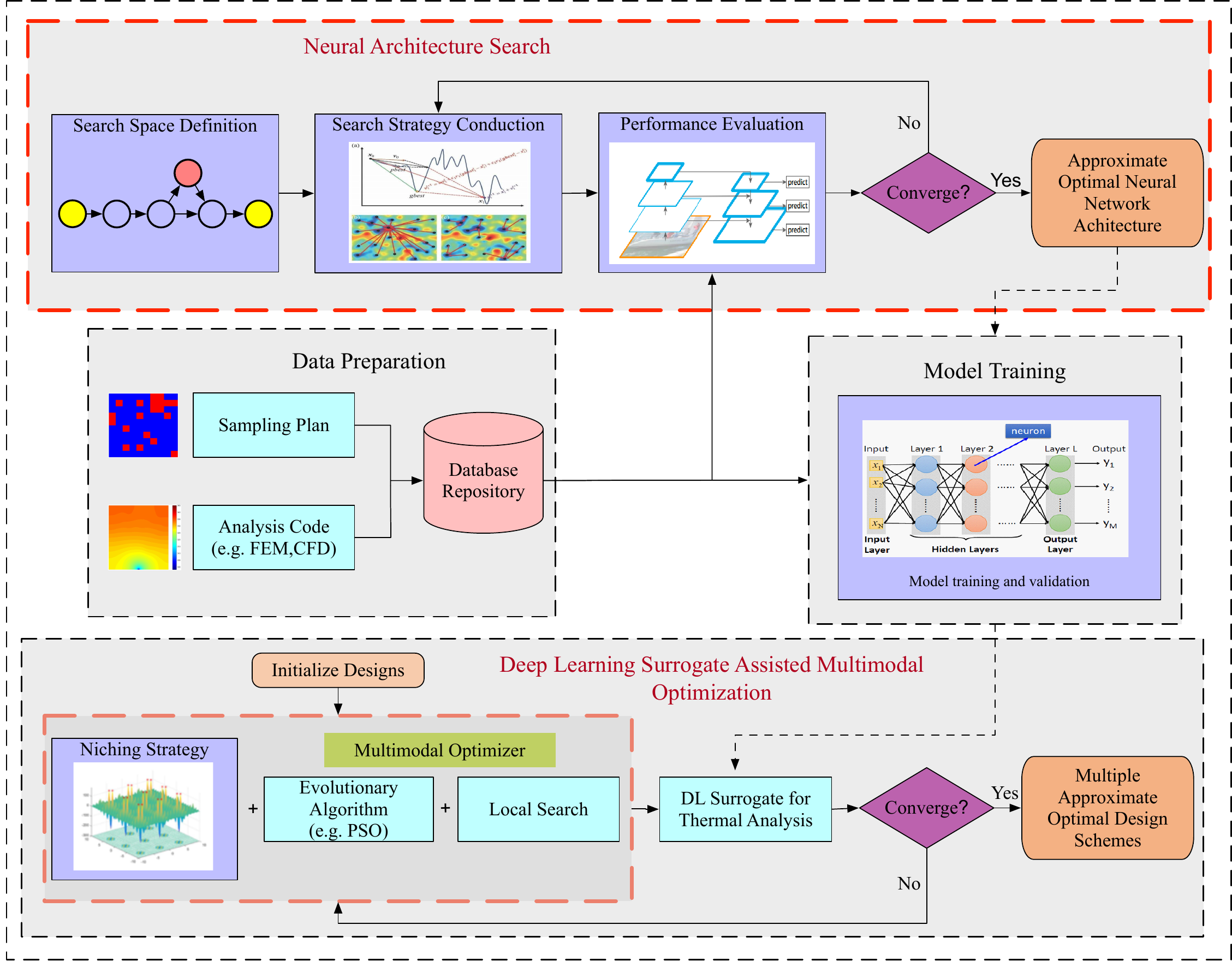}
	\caption{The illustration of our proposed MHSLO-NAS framework.}
	\label{psoms}
\end{figure*}

The deep learning surrogate assisted HSLO includes two core parts: the design of deep learning surrogate and the design of optimization algorithm based on surrogate. On one hand, the deep learning surrogate often needs to be manually designed with rich engineering experience and usually to be complex. On the other hand, the multimodal optimization algorithm which could seek multiple optima simultaneously still need to be researched. To cope with these two difficulties, we propose the framework of multimodal heat source layout optimization design based on neural architecture search (MHSLO-NAS). The brief process of MHSLO-NAS is illustrated in Figure \ref{psoms}.

\iffalse简要精炼的话解释图\fi 

\textbf{Data preparation}. Various samples are generated as the training data and test data according to the specifc sampling strategies. Each sample pair consists of the heat source layout and its corresponding simulated temperature field.

\textbf{Neural architecture search}. Taking the mapping from layout to temperature field as an image-to-image regression task, neural network search is utilized to automatically search for an efficient model architecture with higher accuracy and less parameters. 

\textbf{Model training}. The searched model architecture is retrained from scratch.

\textbf{Deep learning surrogate assisted multimodal optimization} . After the deep learning surrogate model is trained, the multimodal optimization algorithms can be combined to solve HSLO, which could obtain multiple near approximate solutions finally.

\section{Multi-objective neural architecture search for an image-to-image regression task}\label{sec5}

In this section, taking the mapping from layouts to temperature fields as the image-to-image regression task, we introduce the method of using  neural architecture search to obtain the near optimal architecture of backbone in FPN.

\subsection{Search space}\label{search}

\begin{figure}[H]
	\centering
	\includegraphics[scale=0.5]{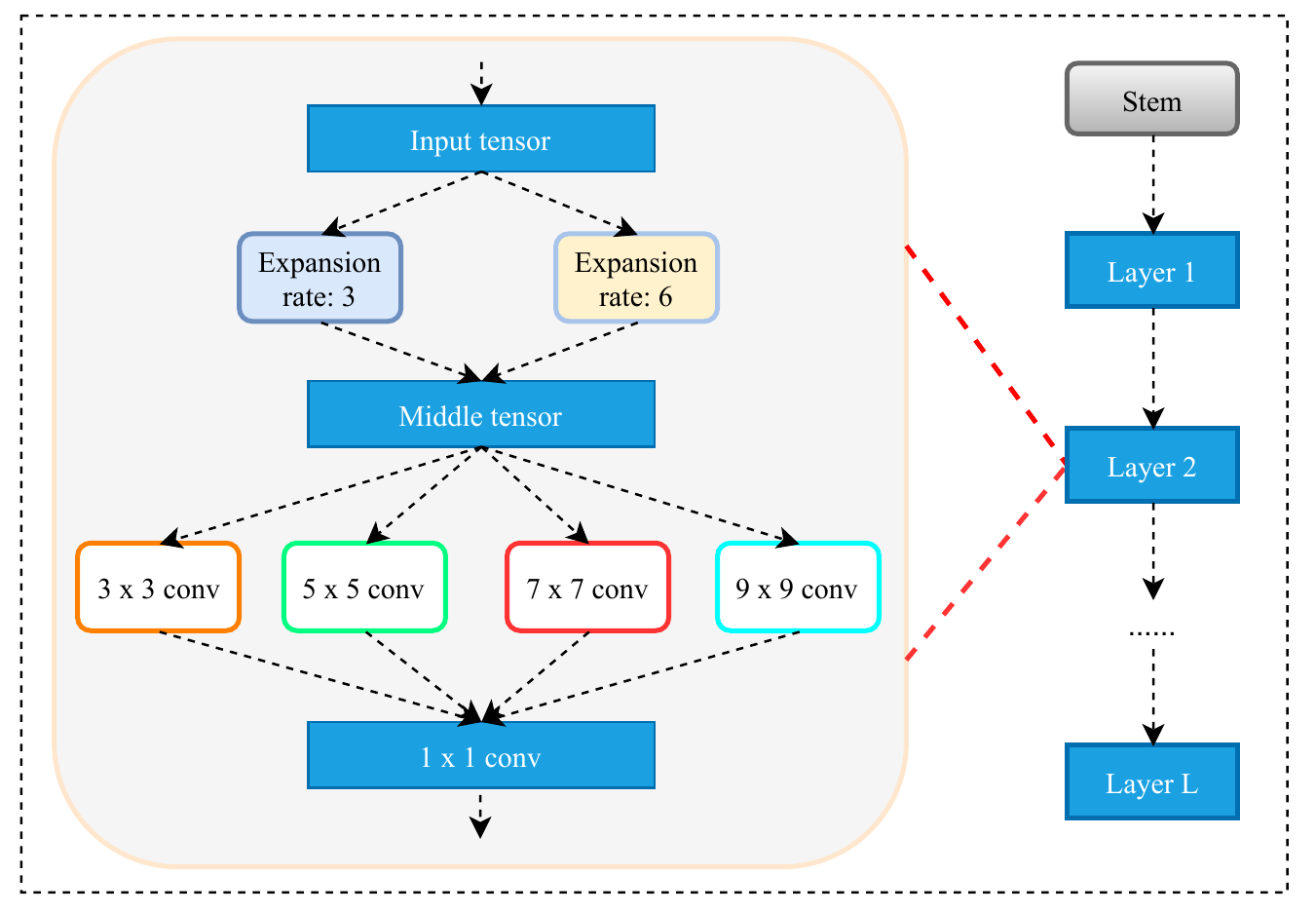}
	\caption{The illustration of the defined search space of the backbone network of Mixpath$\_$FPN model \cite{9368601}.}
	\label{mixpath}
\end{figure}

In original FPN, the backbone is ResNet50 that possesses too large parameters, which brings more training burden. Thus we need to define the sutiable search space to substitute the backbone. Our defining the search space is motivated by MobileNetV2 \cite{Mark2018}, which is an efficient deep learning model by stacking multiple inverted blocks. The structure of the inverted block is visualized in the left part of Figure \ref{mixpath}. Given a pre-defined channel, the real channel in the process of convolutional operation is multiplied by the preset expansion rate in each layer. The size of convolutional kernel in each layer could be selected from 3$\times$3, 5$\times$5, 7$\times$7 and 9$\times$9, which possess various abilities of extracting the feature map. Then the channel of output feature map in each layer would decrease to the pre-defined one by following a conv 1$\times$1 operation that could change the channel of feature map.

Our method trys to search for the near optimal configuration of the convolutional kernel size and expansion rate in a neural network with the fixed length, which is presented in Figure \ref{mixpath}. In each layer, the expansion rate can be selected from [3, 6], while the selection of convolutional kernel is allowed to be multi-path from [3,5,7,9]. $m$ stands for the maximum number of choice paths, which is an random integer belonging to [1, 4]. The input layout images are first fed into a fixed stem layer, which is consisted with C1 in Figure \ref{fpn}. In this paper, we set the number of toal layer as 12. The whole backbone model is divided into four parts uniformly, which are corresponded to C2, C3, C4 and C5 respectively in Figure \ref{fpn}. Then the backbone would be combined with the FPN framwork to form the final Mixpath$\_$FPN model.

\subsection{Search Strategy}\label{strategy}

From the above defined search space, there exist $20^{12}$ configurations. To conduct search srategy in such a large space, a two-stage strategy is utilized as follows.

\textbf{Step 1}: Training the Mixpath$\_$FPN supernet.

\textbf{Step 2}: Conducting multi-objective search based on the trained supernet.

\begin{algorithm}
	\caption{\textbf{Step 1} Training the Mixpath$\_$FPN Supernet}
	\label{alg1}
		\begin{algorithmic}[1]
	\Require The supernet with $L$ layers, the maximum optional paths $m$, the epoch of training the neural network $N$, the parameter of the supernet $\Theta$, dataset $D$ consisting of various layout schemes and temperature fields.
	\Ensure The trained Mixpath$\_$FPN supernet including all choice paths.
	
			\For{$i \leftarrow 1:N$}
			\For{$i \leftarrow 1:L$}
				\State	$r$ $\leftarrow $ Select a value from [3, 6] as the expansion rate randomly;
				\State	$m^{\prime}$ $\leftarrow $ Select a value from [0, $m$] randomly as the number of choice paths;
				\State	Select $m^{\prime}$ values from [3, 5, 7, 9] randomly without repeating as the configuration of convolutional kernel size; 
		\State	Obtain the sub-model from the supernet using above sampled configuration;
		\State	Calculate the gradients based on dataset $D$ and update the parameter of supernet $\Theta$ ;
	\EndFor
	\EndFor
	\end{algorithmic}
\end{algorithm}

\begin{algorithm}
	
	\caption{\textbf{Step 2} Conducting multi-objective search based on the trained supernet}
	\label{alg2}
	\begin{algorithmic}[1]
	\Require The supernet $\bm{S}$, the maximum number of iteration ${N}$, population size $n$, validation set $D$, the crossover rate $P_{c}$, the mutation rate $P_{m}$.
	\Ensure The near optimal model architectures on the Pareto front.
	 \State  Generate the initial populations $\bm{P_{1}}$  with $n$ candidate architectures randomly based on $\bm{S}$;
	
		\For{$i \leftarrow 1:N$}
			\State $\bm {{Q_{i}}}$ = $ \emptyset $;
			\For{$j \leftarrow 0:2:n-1$}
				
			\State	${q_{j+1}}$ = crossover($\bm{P_{i}}$, $P_{c}$);
			\State	${q_{j+2}}$ = mutation($\bm{P_{i}}$, $P_{m}$);
			\State Add ${q_{j+1}}$ and ${q_{j+2}}$ to  $ \bm {Q}$;
			\EndFor
		 \State	Merge the parent and children population together: $ \bm{R_{i}}$$=\bm {P_{i}} \cup \bm {Q_{i}}$ ;
			\State Calculate the prediction accuracy and total parameter of each architecture in $\bm{R_{i}}$ using $\bm{S}$ in $D$;
			\State Calculate the Pareto frontier: $F$=non-dominated-sorting($\bm{R_{i}}$)			;
		\State	$\bm {P_{i+1}}$ $\leftarrow$ Select $n$ architectures to obtain $\bm {P_{i+1}}$ according to the crowding distance: ${P_{i+1}}$ = selection($\bm{R_{i}}$)	;
			
		\EndFor
	\end{algorithmic}
\end{algorithm}

The core idea of above steps is first training a supernet including all possible paths. During the training process, each path would be selected randomly and the correponding sub-model is trained. After the training process terminates, even though the trained supernet could not reach the high accuracy of directly being utilized to predict the temperature field, it has the certain ranking ability to evaluate the comparative performance of different sub-models. Then, evolutionary algorithms could be easily combined to find the near optimal model architecture on the basis of the trained supernet. The searched model would be reatrained from scratch to meet the requirement of high enough accuracy to predict the temperature field.

In our solving process, we need to encode the individual firstly. One of  choice paths could be taken as an example to be interpreted,  which is encoded as follows:
\begin{equation}
\left\{\begin{array}{ll}
$0: {'$conv$': [3, \ 5], \qquad  \ \ '$rate$': 3},$\\
$1: {'$conv$': [3, \ 7, \ 9],  \quad \ '$rate$': 6},$\\
...\\

$11: {'$conv$': [5, \ 7, \ 9], \quad '$rate$': 3},$\\
\end{array}\right.
\label{equn}
\end{equation}

In detail, the process of training the supernet is presented in \textbf{Algorithm \ref{alg1}}. First, the supernet including all choice paths is built. Then the defined heat source layout dataset are utilized to train the supernet. Different from training the whole neural network, we train the sub-model by choosing the random path such as Eq. (6) at each time. The configurations of expansion rate and kernel size in each layer are all randomly selected during the training. After training the supernet for $N$ epochs, we adopt \textbf{Algorithm \ref{alg2}} to search for the near optimal sub-model. Taking both of the model parameters and predicted accuracy into consideration, we model it as a two-objective optimization problem. Thus, multi-objective evolutionary algorithm can be performed to solve it. In the crossover operation, after two individual in $\bm{P_{i}}$ are selected randomly, their configuration in each layer are exchanged randomly with the probability of $P_{c}$. In the mutation operation, after an individual in $\bm{P_{i}}$ is ranomly selected, we refine the configuration randomly according with the probability of $P_{m}$. The generated individuals are all saved to $\bm Q$. After the population $\bm{P_{i}}$ and $\bm{Q_{i}}$ are merged, we obtain the solutions $F$ by non-dominated sorting. All of the individuals in $F$ are regarded as the near optimal model architetctures, which meets our requirements for less parameters or higher prediction accuracy.

\subsection{Performance Evaluation}\label{sec41}
The prediction of deep learning surrogate is a temperature field with 200 $\times$ 200 grid.  In the training process, the MAE between the prediction and ground-truth is selected as the loss function. The absolute error (AE) between the predicted value $\hat{y}$ and the ground-truth value $y$ is
defined as
\begin{equation}
\operatorname{AE}(\hat{y}, y)= \lvert (\hat{y}-y) \rvert
\end{equation}
then the mean absolute error (MAE) between the predicted temperature matrix $\boldsymbol{Y}$ and the ground-truth temperature matrix $\widehat{\boldsymbol{Y}}$ is defined as
\begin{equation}
\operatorname{MAE}(\widehat{\boldsymbol{Y}}, \boldsymbol{Y})=\frac{1}{40000} \sum_{i=1}^{200} \sum_{j=1}^{200} \mathrm{AE}\left(\widehat{\boldsymbol{Y}}_{i, j}, \boldsymbol{Y}_{i, j}\right) \
\end{equation}

We evaluate the performance of the searched model and other models from the following four aspects: 

\textbf{Accuracy}: Lower MAE means higher prediction accuracy. 

\textbf{Parameters}: The total parameters of the neural network. 

\textbf{FLOP}$_{s}$\cite{flop}: The number of floating-point operations, which could assess the time complexity of the neural network. 

\textbf{Inference time}: The time of executing one forward calculation of neural network on average.

\section{Multimodal neighborhood search based layout optimization algorithm}\label{sec6}
Based on the description in Section \ref{sec2}, we can model the heat source layout optimization as a discrete interger optimization problem. The layout domain is divided into 10 x 10 grid. The input design space could be represented by a 20-dimensional vector. Each element of the vector stands for the position of a heat source in the layout domain. Thus the value of each element ranges from 1 to 100. Taking the layout scheme shown in Figure \ref{vp}(b) as an example, the layout could be represented by a sequence as following:

\begin{equation}
\begin{aligned}
\bm {X} = &[3, 4, 5, 9, 16, 15, 17, 59, 61, 64, 66, 70, 72 ,\\
& 78, 84, 89, 91, 94, 100]
\end{aligned}
\end{equation}
When a layout scheme is given, the corresponding temperature field is calculated by the searched neural network other than simulation tool. So the whole mathemetical model could be illustrated as follows:

\begin{equation}
\left\{\begin{array}{ll}
\text {find } & \boldsymbol{X}=\left\{x_{i}, i=1,2, \ldots, 20\right\} \\
\text {minimize } & \widehat{R}_{m}=\hat{f}(\boldsymbol{X}) \\
\text {s.t.} & 1 \leq x_{i} \leq 100  \& x_{i} \in N \quad \forall i \\
& x_{i} \neq x_{j} \quad \forall i \neq j
\end{array}\right.
\label{equn}
\end{equation}
where $\widehat{R}_{m}$ stands for the predicted maximum temperature of the layout domain calculated by the searched  Mixpath$\_$FPN surrogate model, the surrogate model is denoted by $\hat{f}$.

Then we demonstrate the algorithm process of searching multiple optimal solutions by use of the searched Mixpath$\_$FPN surrogate model instead of a large number of time-consuming heat simulation. To realize it, a multimodal neighborhood search based layout optimization algorithm (MNSLO) is developed as the optimizer to solve above-mentioned HSLO problem, which is provided in \textbf{Algorithm \ref{alg:3}}.

\begin{algorithm}
	\caption{Multimodal  neighborhood Search  based Layout Optimization Algorithm  (MNSLO)}
	\label{alg:3}
	\begin{algorithmic}[1]
		\Require The population size $NP$, the layout solution $\bm{X} = \{x_{j}, j = 1,2,...,20\}$, the number of optimal solutions set $n$, the level to select multiple solutions 	$\epsilon$, deep learning surrogate $\hat{f}$, the number of groups $c$
		\Ensure The set of optimal candidate layout solutions $\bm {S} $
		\State Randomly initialize the initial population $\bm{X_{0}}={x_{i0},i=1,2,...,NP}$ with $NP$ individuals; 
		\State Use the Mixapath$\_${FPN} surrogate to calculate the fitness value $\bm {\hat{f}(X_{0}) }$;
		\State Initialize the optimal solution set $\bm {S} $;
		\State Divide the whole population into $c$ groups according to \textbf{Algorithm} \ref{alg:4};
		\For{$each$ $leader$ $individual$ $in$ $each$ $group$}
		\State	$X_{g}$ $\leftarrow$ the leader individual; 
		\While{$flag = 1$}
		\State	$flag = 0$;
		\State Randomly generate an integer sequence $\bm R$$ =\{r_{j}, j = 1,2,...,20, 1 \leq r_{j} \leq 20\}$ without repeating; 
		\For{$j \leftarrow 1:20$}
		\State	$i $$\leftarrow$  $r_{j}$, determine the position of generating the neighborhood solutions; 
		\State	Generate the set of neighbour candidate solutions: $N(X_{g},i)=neighborhood(X_{g},i)$ according to \textbf{Algorithm} \ref{alg:5};
		\State	Calculate the fitness values of neighborhood solutions $\hat{f}(N(X_{g},i))$;
		\For{each $X\in N{(X_{g},i)}$}
		\If{$\hat{f}(X) $$< min_{X\in N{(X_{g},i)}}$$\hat{f}(X) + \epsilon$}
		\State 	Add $X$ to $\bm {S} $;
		\EndIf
		\EndFor
		\State $\bm {S} $	 $\leftarrow$	Select $n$ optimal solutions from $\bm {S} $;
		\If{ $min_{X\in N{(X_{g},i)}}$$\hat{f}(X)<fitness_{g}$}
		\State	$flag = 1$;
		\State	$fitness_{g}$ $\leftarrow$ $min_{X\in N{(X_{g},i)}}$;
		\State	$X_{g}$ $\leftarrow$ $argmin_{X\in N{(X_{g},i)}}$;
		\EndIf
		\EndFor
		\EndWhile
		\EndFor
	\end{algorithmic}
\end{algorithm}

With regards to the multimodal optimization problem, maintaining the diversity of population plays a key role. To preserve the population diversity, we utilize the clustering algorithm according to the similarity of different individuals, which is illustrated in \textbf{Algorithm \ref{alg:4}}. In the clustering algorithm, we first need to calculate the pairwise similarity between two individuals. In this study, the similarity of two permutations $\pi_{i}$ and $\pi_{j}$ is defined as follows:


\begin{equation}
s\left(\pi_{i}, \pi_{j}\right)=\frac{\sum_{d=1}^{20} \lvert(\pi_{i,d} - \pi_{j,d}\rvert)}{N}
\label{eq12}
\end{equation}
where $\pi_{i}$ denotes the sequence of $i_{th}$ individuals standing for the heat source layout, $N$ is the number of heat source. In detail, the objective of clustering at first is to gather the similar individuals into one group. After calculating the fitness values of the population $\bm{P}$ by deep learning surrogate $\hat{f}$, the best individual is selected as the leader. Then the similarities with it of all other individuals are calculated and sorted. Other individuals are combined with the best individual as a group until the size of group is meeted.

\begin{algorithm}
	\caption{Clustering Algorithm}
	\label{alg:4}
	\begin{algorithmic}[1]
	\Require The population $\bm{P}$ with $NP$ individuals, the number of groups $c$, deep learning surrogate $\hat{f}$.
	\Ensure A set of groups.
	\State Determine the cluster size $M = NP/c $;
	\State Sort $\bm{P}$ according to the fitness value calculated by $\hat{f}$ in descend order;
	\While {size of $\bm{P} > 0 $}
		\State Select the best individual $\bm{P}_{best}$ in $\bm{P}$ as the leader individual;
		\State Cluster the $M-1$ individuals nearest to  $\bm{P}_{best}$ and $\bm{P}_{best}$ as a group according to Eq.(\ref{eq12});
		\State Delete these  $M$ individuals from $\bm{P}$;	
	\EndWhile
	
			\end{algorithmic}
\end{algorithm}

\begin{algorithm*}[h]
	\caption{Generate the $t$th neighbour  layout solution $X$:$ N(X,t)=neighborhood(X,t)$}
	\label{alg:5}
\begin{algorithmic}[1]
	\Require The layout solution $\bm{X} = \{x_{j}, j = 1,2,...,20\}$.
	\Ensure The set of neighbour candidate solutions $ N(X,t)$.
\State 	Randomly generate an integer sequence $\bm R$$ =\{r_{j}, j = 1,2,...,100, 1 \leq r_{j} \leq 100\}$ without repeating;
	\For{$j \leftarrow 1:100$}
	\State	$i $$\leftarrow$  $r_{j}$, determine the position of generating the new solution; 
	\State	$ X_{neighbor} = X$
		\If {$i \notin X$} 
		\State	$x_{t}$ $\leftarrow$ $i$ where  $x_{t}$ is the $t_{th}$ position number in $X_{neighbor}$;
		\EndIf
		 
		\State	$k$ $\leftarrow$ find the position number: $x_{k}=i$ in $X_{neighbor}$; 
		\State	$temp$ $\leftarrow$ $x_{t}$;
		\State	$x_{t}$ $\leftarrow$ $i$;
		\State	$x_{k}$ $\leftarrow$ $temp$;
		
		\State Include the new $X_{neighbor}$ in $N(X, t)$;
	\EndFor
		\end{algorithmic}
\end{algorithm*}

By clustering the whole population into multiple groups, we could obtain the intial layout as diversely as possible. Then we conduct neighborhood local search based on the leader individual to improve the global searching ability of the algorithm. Different from the NSLO in \cite{Chen2020}, we improve it from four aspects in local search. First, the position of heat source to conduct local search is selected randomly. So we generate the sequence $\bm R$$ =\{r_{j}, j = 1,2,...,20, 1 \leq r_{j} \leq 20\}$ randomly. Second, after selecting the position of conducting local search, we generate the neighborhood candidate solutions by moving a heat source each time according to \textbf{Algorithm \ref{alg:5}}. We define the number of neighborhood candidate solutions to 99. Apart from detecting the positions of not being layouted, we also exchange the two positions of having being layouted. So our strategy could possess stronger adaptivity in more complex layout problem. In this process, we also randomly generate the sequence to determine which position first to be compared. To realize the purpose of multimodal optimization, we define an optimal solution set $S$ with the fixed size. In the local search, we would preserve multiple superior solutions into $S$. A threshold $\epsilon$ is utilized to select the near optimal solutions. However, in every iteration, a selection operation is conducted on $S$ to preserve $n$ optimal solutions.\\

\section{Experiments result}\label{sec7}

In this section, two examples are utilized to demonstrate and verify the effectiveness of the proposed method. Case 1 is the same as Chen et.al \cite{Chen2020}. Case 2 demonstrates a 20 heat source layout optimization problem with  different heat intensities. In these two cases, we evaluate the proposed method from two aspects:

1. On one hand, we evaluate the performance of deep learning surrogate model searched by multi-objective neural architecture search method on the test set:
\begin{itemize}
	\item  Whether the search strategy on the search space is better than random search.
	
	\item How well does the searched model perform on the test set compared with previous FPN.
	
\end{itemize}

2. On the other hand, we evaluate the performance of the proposed multimodal discrete evolutionary algorithm:

\begin{itemize}
	\item Whether the searched best layout scheme is better compared with other optimization method.
	
	\item Whether the algorithm could seek multiple optima to provide more layout design choices.
	
\end{itemize}

\begin{table}[h]
	\renewcommand\arraystretch{1}
	\centering
	\caption{The statistic of training and test data.}
	\setlength{\tabcolsep}{8mm}{
		\begin{tabular}{ccc}
			\toprule
			
			{Data Type} & {Training} & {Test}  \\ \midrule
			{Size} & 30,000   & 5,000  \\
			\bottomrule
	\end{tabular}}
	
	\label{tab1}
\end{table}

In our experiments, the detail settings are as follows. In the process of NAS, the training epoch, learning rate and batch size of the Mixpath$\_$FPN supernet are set to 600, ${10}^{-3}$, 32 respectively. In the process of NSGA-II for searching architectures, the population size, $P_{c}$ and $P_{m}$ are set to 40, 1, 1. After the final model according to the Pareto front of NSGA-II based on the supernet is obtained, we retrain the neural network from scratch. The epoch of retraining is set to 50. The preset channel of each layer in the supernet is set to [32, 48, 48, 96, 96, 96, 192, 192, 192, 256, 256, 320, 320]. When the multimodal optimization is conducted based on the deep learning surrogate, the population size is set to 30. All the experiments are implemented under the same computational experiment: Intel(R) Core(TM) i7-8700 CPU @3.2GHz and 1 NVIDIA Tsela P100 GPU with 16G memory.

The data is shown in Table~\ref{tab1} in detail. 30,000 training samples are generated randomly, and 5000 test samples are generated randomly for testing the performance of trained deep learning model.

\subsection{Case 1:  heat source layout optimization with the same intensity}

In this case, the parameters of heat source are set according to Chen et.al \cite{Chen2020}, which has been introduced in Section \ref{sec2b}. Detailed sampling strategy to generate the training data and test data could be seen in \cite{Chen2020}.

\subsubsection{The performance of the searched model}

\begin{figure}[h]
	\centering
	\includegraphics[scale=0.45]{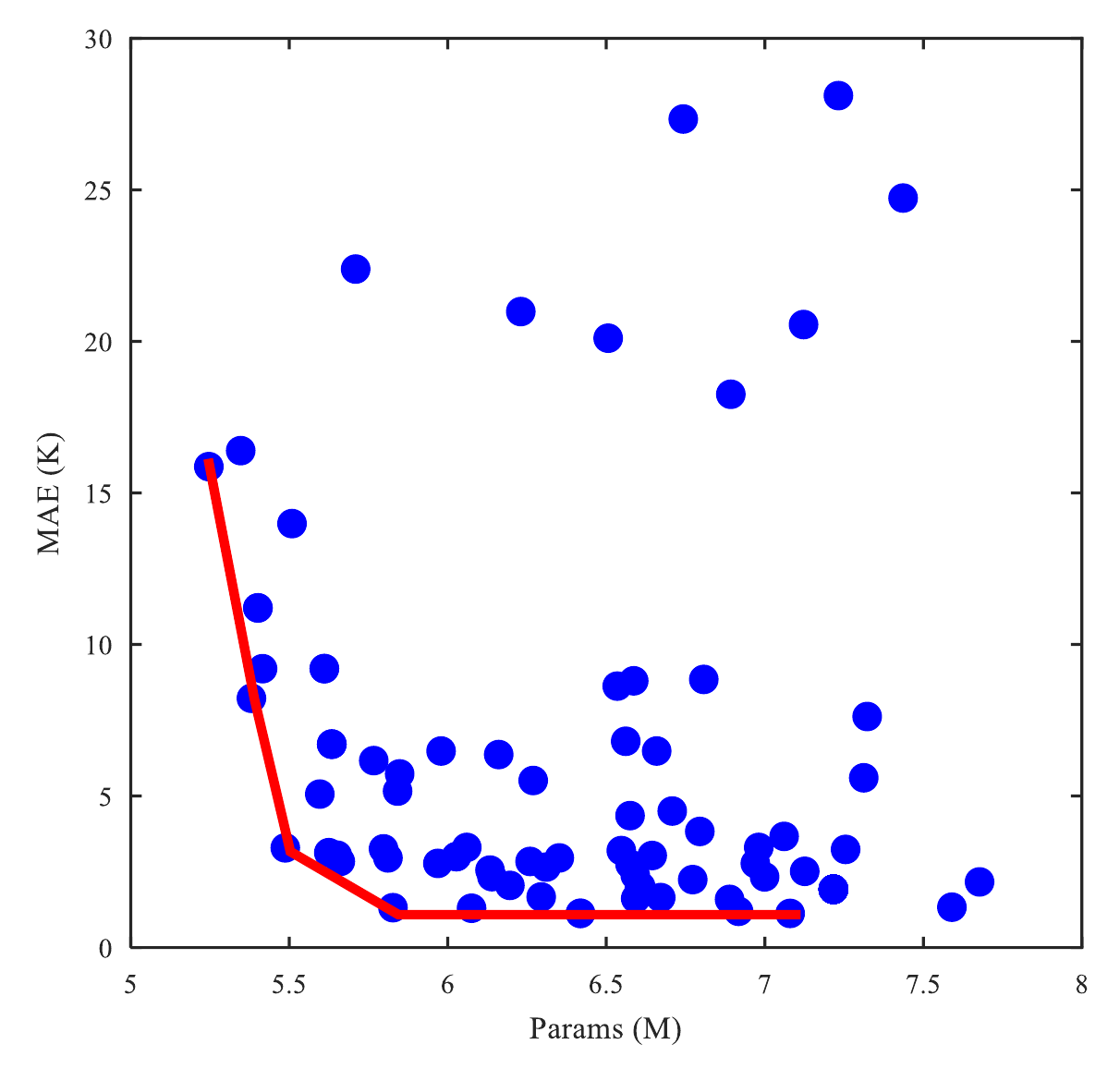}
	\caption{The pareto front of the searched models by NSGA-II in case 1.}
	\label{nsga1}
\end{figure}

\begin{figure}[h]
	\centering
	\includegraphics[scale=0.5]{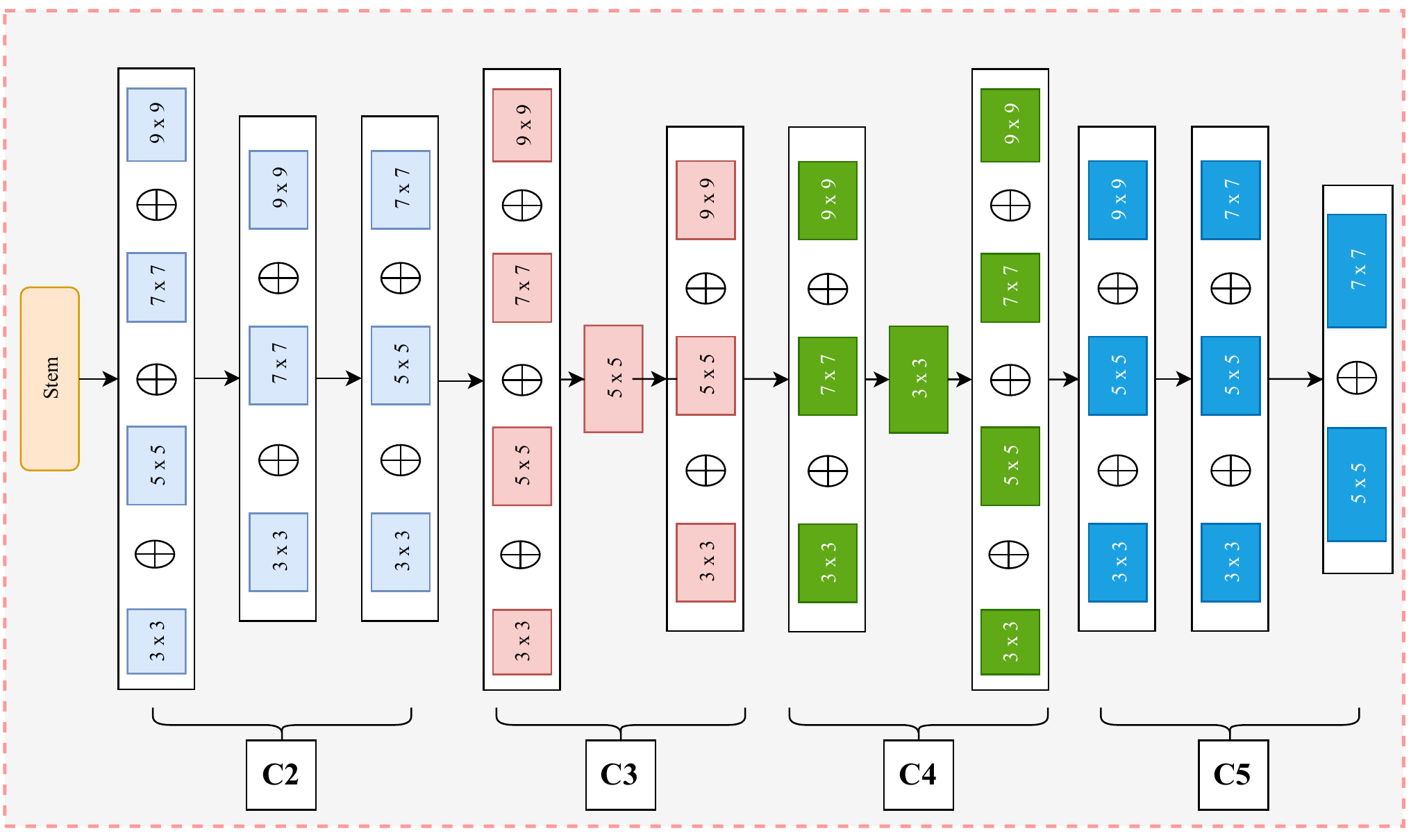}
	\caption{The visualization of the searched backbone architecture in case 1}
	\label{archi}
\end{figure}

After training the Mixpath$\_$FPN supernet for 200 epochs, we search for the optimal architecture using NSGA-II algorithm. The result optimized by NSGA-II in case 1 is presented in Figure \ref{nsga1}. Each blue dot represents a kind of model architecture. We also show the Pareto frontier, which is the non-dominated solution set. Designers could select the suitable model according to the practical need. For example, though the accuracy of models with less parameters  decreases to some extent, it could have faster inference time. In case 1, we select one model architecture comprehensively from the Pareto frontier. The selected architecture is presented in Figure \ref{archi}. We evaluate the performance of our searched model from four criteria introduced in Section \ref{sec41}.

%
%
%

\begin{figure}[h]
	\centering
	\includegraphics[scale=1]{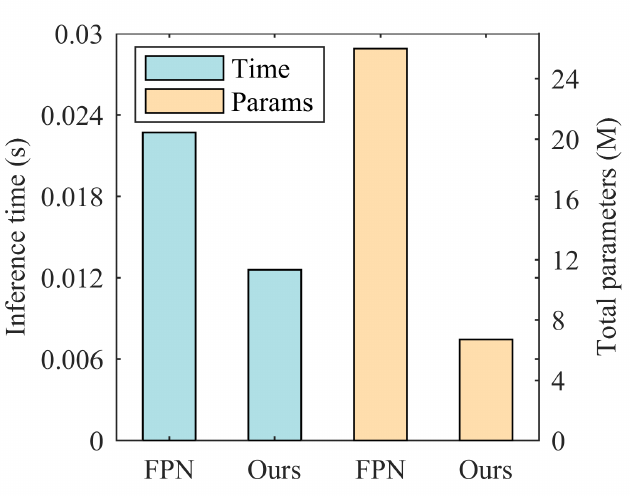}
	\caption{The comparison of the performance of the searched model and original FPN in case 1.}
	\label{time2}
\end{figure}

\begin{table}[htb]
	\scriptsize
	
	\centering
	
	\caption{The comparison of the searched model by our developed NAS method with other models on two cases. FDM (the second row): the time cost of calculating the thermal performance of one heat source layout scheme using traditional FEM method. FPN : the result of classical FPN model with ResNet50 as the backbone. Mixpath$\_$FPN$\_${i} (i = $random, small, large$): the result of the  mannually designed model. Mixpath$\_$FPN: the result of the searched model by our method.}
	
	\label{Tab03}
	\setlength{\tabcolsep}{1.5mm}
	{
		\begin{tabular}{ccccccccccc}
			
			\toprule
			
			$Case$ & $Model$	&  $Accuracy$ 	&  $Params$   &   $Flops$  & $Inference \ time$		   \\		
			\midrule		
			\multirow{12}{*}{Case 1}& {FDM}   &    -     &     -       & -       &    0.3034s   \\				
			\cline{2-6}
			& {FPN \cite{Chen2020}}   & 0.108K         & 26M           & 4.97G       &    0.0228s   \\		
			& {ResNet18$\_$FPN} \cite{7780459}  & 0.126K         & 13.04M           & 2.77G       &    0.0203s   \\		
			& {DetNet$\_$FPN \cite{DBLP:conf/eccv/LiPYZDS18}}   & 0.108K         & 20.68M           & 1.67G       &    0.0658s   \\
			& {MobileNet$\_$FPN \cite{DBLP:journals/corr/HowardZCKWWAA17}}   & 3.801K        & 3.11M           & 1.10G       &    0.0807s   \\
			& {Unet \cite{DBLP:conf/miccai/RonnebergerFB15}}   & 0.123K         & 22.93M           & 14.56G       &    0.0173s   \\			
			& {Unet (scale=0.5) \cite{DBLP:conf/miccai/RonnebergerFB15}}   & 0.123K         & 6.08M           & 4.26G       &    0.0173s   \\
			& {ResNet34$\_$Unet \cite{7780459}}   & 0.104K        & 103.08M           & 21.05G       &    0.0828s   \\
			\cline{2-6}
			& \text { Mixpath$\_$FPN$\_random$ }   &0.153K       & 6.49M & 1.44G &  0.0124s\\
			& 	\text { Mixpath$\_$FPN$\_small$ } &  0.193K   & 4.75M          & 1.22G       &      0.0113s\\
			&	\text{ Mixpath$\_$FPN$\_large$ }   & 0.135K  & 8.97M           & 2.42G          &  0.0144s  \\
			
			\cline{2-6}
			&\text { Mixpath$\_$FPN (Ours)}   &\textbf{0.105K}       & \textbf{6.57M}  & \textbf{1.81G}  &\textbf{0.0124s}\\
			
						\hline
			\multirow{11}{*}{Case 2}& {FPN \cite{Chen2020}}   & 0.069K         & 26M           & 4.97G    &       0.0248s     \\	
			& {ResNet18$\_$FPN} \cite{7780459}  & 0.119k         & 13.04M           & 2.77G       &    0.0209s   \\		
			& {DetNet$\_$FPN \cite{DBLP:conf/eccv/LiPYZDS18}}  & 0.281K        & 20.68M           & 1.67G       &    0.0648s   \\
			& {MobileNet$\_$FPN \cite{DBLP:journals/corr/HowardZCKWWAA17}}  & 13.26k         & 3.11M           & 1.10G       &    0.0704s   \\
			& {Unet \cite{DBLP:conf/miccai/RonnebergerFB15}}    & 0.159K         & 22.93M           & 14.56G       &    0.0173s   \\			
			& {Unet (scale=0.5) \cite{DBLP:conf/miccai/RonnebergerFB15}}   & 0.123K         & 6.08M           & 4.26G       &    0.0173s   \\
			& {ResNet34$\_$Unet \cite{7780459}}    & 0.253K         & 103.08M           & 21.05G        &    0.0828s   \\
			\cline{2-6}
			& \text { Mixpath$\_$FPN$\_random$ }   &0.139K      & 6.87M  & 1.91G &  0.0124s\\
			& 	\text { Mixpath$\_$FPN$\_small$ } &  0.187K   & 4.75M          & 1.22G      & 0.0113s     \\
			&	\text{ Mixpath$\_$FPN$\_large$ }   & 0.125K  & 8.97M           & 2.42G      &  0.0154s      \\
			\cline{2-6}
			&\text { Mixpath$\_$FPN (Ours) }   &\textbf{0.095K}       & \textbf{6.17M}  & \textbf{1.73G} &\textbf{0.0146s} \\

			\bottomrule		
	\end{tabular}}	
\end{table}

To evaluate the effect of the search strategy, we generate three models. The model with the least paths is denoted as Mixpath$\_$FPN$\_small$. In this model, each layer only has a kind of kernel size 3x3. The expansion rate of each layer is set to 3. The model with the most paths is denoted as Mixpath$\_$FPN$\_large$.  In this model, each layer only has four kinds of kernel sizes. The expansion rate of each layer is set to 6. The model with a random path is denoted as Mixpath$\_$FPN$\_random$. The result of these three models is shown in the fourth row to the sixth row in Table~\ref{Tab03}. We could see that the accuracy of our searched model reach 0.105K, which is higher than all other models. Compared with the mannually designed model, the result proves that our NSGA-II search strategy seeks the better model architecture. 

\begin{table}[h]
	\centering
	\caption{The statistic of total inference time of different sample size tested on the original FPN model and the searched Mixpath$\_$FPN model. (time unit: s)}
	\setlength{\tabcolsep}{3mm}{
		\begin{tabular}{ccccc}
			\toprule
			
			{ Model} & {10000} & 20000 & 30000 & 40000 \\ \midrule
			{\sc \ FPN} & 147.12 & 294.56    & 442.15  & 589.13\\
			{ Mixpath$\_$FPN} & \textbf{97.86}   & \textbf{203.05}  & \textbf{302.32} & \textbf{395.72} \\
			\bottomrule
	\end{tabular}}
	
	\label{totaltime}
\end{table}

To illustrate the effect of the searched model by NAS compared with the handcrfated models, we assess 7 popular models in image segmentation and the searched Mixpath$\_$FPN model on the test data using four metrics introduced in Section~\ref{sec41}. These 7 models include the original FPN with ResNet50 \cite{Chen2020}, ResNet18$\_$FPN \cite{7780459}, DetNet$\_$FPN \cite{DBLP:conf/eccv/LiPYZDS18}, MobileNet$\_$FPN \cite{DBLP:journals/corr/HowardZCKWWAA17}, Unet \cite{DBLP:conf/miccai/RonnebergerFB15}, Unet (scale=0.5) \cite{DBLP:conf/miccai/RonnebergerFB15} and ResNet34$\_$Unet \cite{7780459} respectively. The reason of choosing these models is that all of them achieve remarkable performance on image classification or segmentation. Because the model with the encoder-to-decoder structure is suitable for the iamge-to-image regression task, we choose the FPN, Unet and their variants as a comparison. Among them, MobileNet is a series of efficient model that could be deployed on mobile terminal. The setting of model architectures and channels is the same as \cite{DBLP:journals/corr/HowardZCKWWAA17}. Then the model is combined into FPN framework as the backbone to form  MobileNet$\_$FPN. Other models such as ResNet18 and ResNet34 are also combined into FPN framework as the backbone. The channels of Unet is [64, 128. 256, 512, 1024], which is tha same as the original version reported in  \cite{DBLP:conf/miccai/RonnebergerFB15}. We also implemented it by decreasing the channels to half, [32, 64, 128, 256 ,512], which is denoted as Unet (scale=0.5). To make these models sutiable for the HSLO task, we adjust the resolution different from the original version reported in the literature. The detail setting of resolution is the same as Section \ref{sec2b}. Then we evaluate the performance of them on the same test data on the same computational environment. Their performances on two cases are listed in Table~\ref{Tab03}. It could be seen that with the similar prediction accuracy, our searched model possesses only 1/4 times total parameters than FPN. The total parameters of the original FPN could reach 26M, while that of the searched Mixpath$\_$FPN is only 6.57M. Though the total parameters of MobileNet$\_$FPN  is also only 3.11M, the predicted accuracy is farther lower than other models, which is only 3.801K with MAE. In addition, the FLOP$_{s}$ of the searched model can be reduced from 4.97G to 1.81G compared with the original FPN, which decreases greatly the training cost. We also calculate the average time of simulating one heat source layout by FDM \cite{ReimerA2013} with the grid 200 $\times$ 200 for 1000 times. The average time cost by FDM is around 0.3034s. It could be seen that the original FPN is around 0.0228s on average from Figure \ref{time2}, which proves the effectiveness of decreasing the calculation cost by using deep learning surrogate. However, the result also shows that the average time cost of the searched model by us could further be reduced to 0.0124s, which is 36$\%$ faster than original FPN. Due to the large number of objective function evaluation in HSLO, thus the whole optimization process with the searched model by us would be more efficient. We set the number of calculating the objective function to 10000, 20000, 30000, and 40000 respectively. Then we use FPN model and the searched Mixpath$\_$FPN model to make predictions. The stastic of total inference time is listed in Table \ref{totaltime}. As we can see, the design of the smaller and efficient deep learnig model is helpful to further decrease the computational cost.

\begin{figure*}[htb]
	\centering
	\subfigure 
	{\includegraphics[scale=0.05]{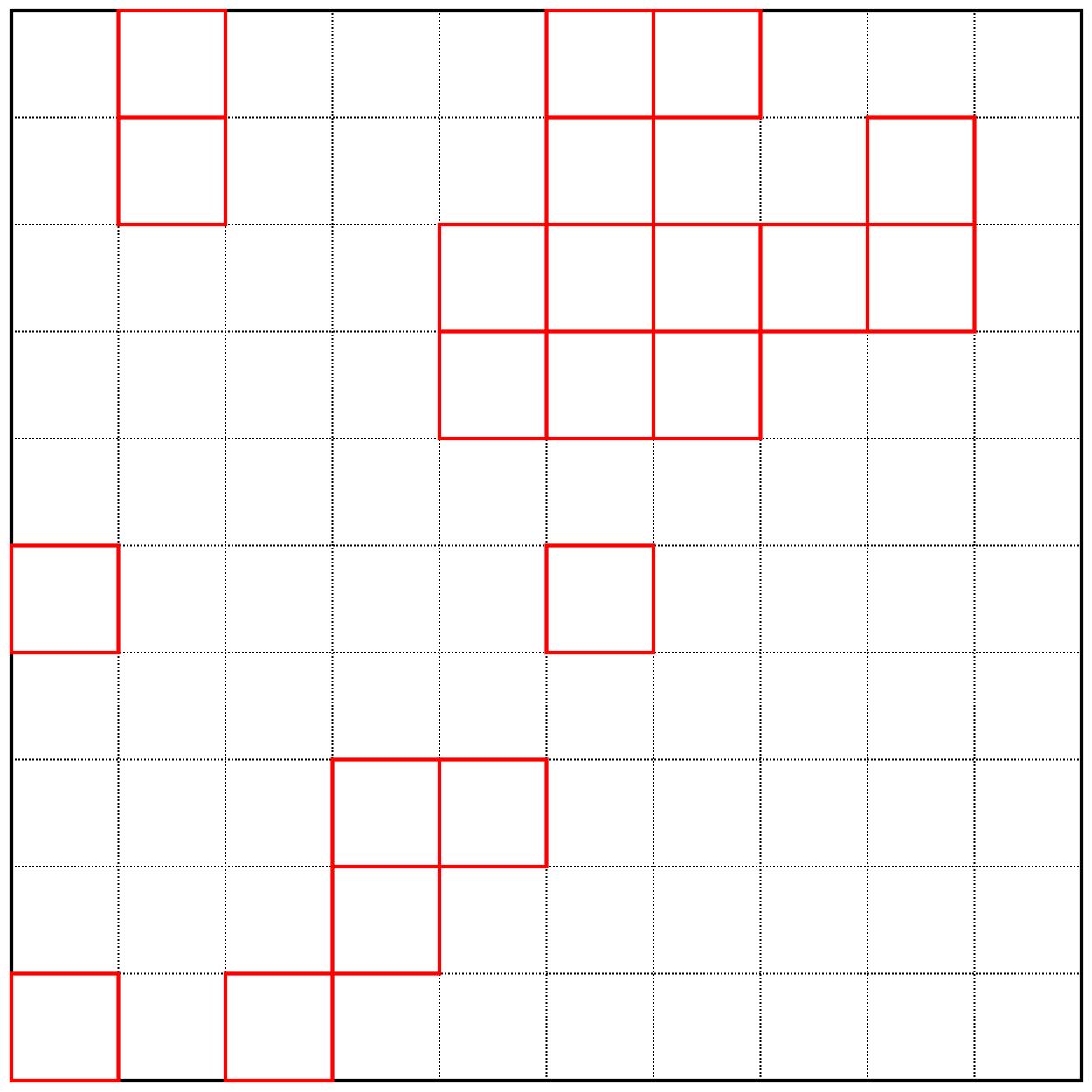}\label{fig9a}} 
	\hspace{3mm}
	\subfigure  
	{\includegraphics[scale=0.05]{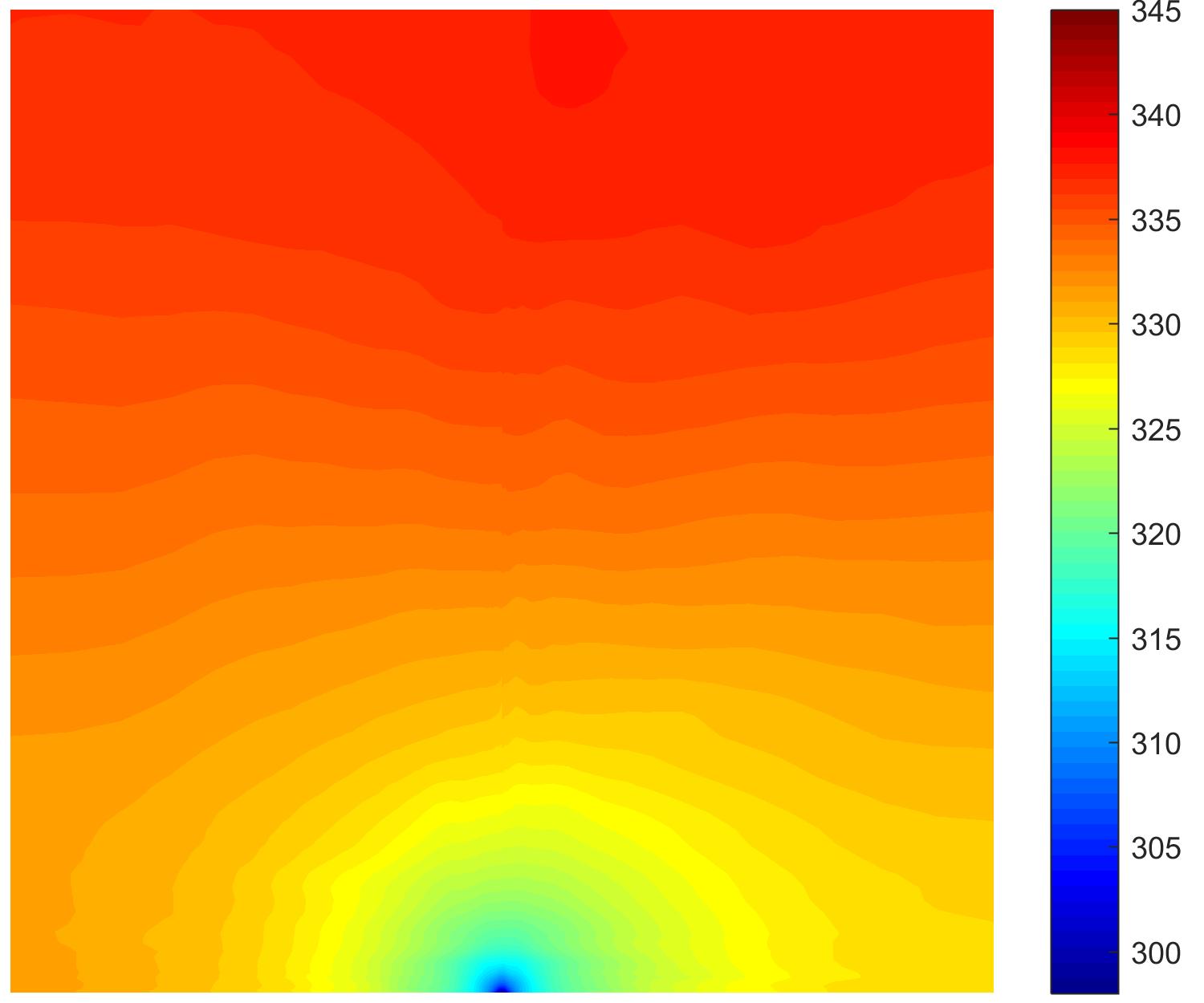}\label{fig9b}}
	\hspace{3mm}
	\subfigure 
	{\includegraphics[scale=0.05]{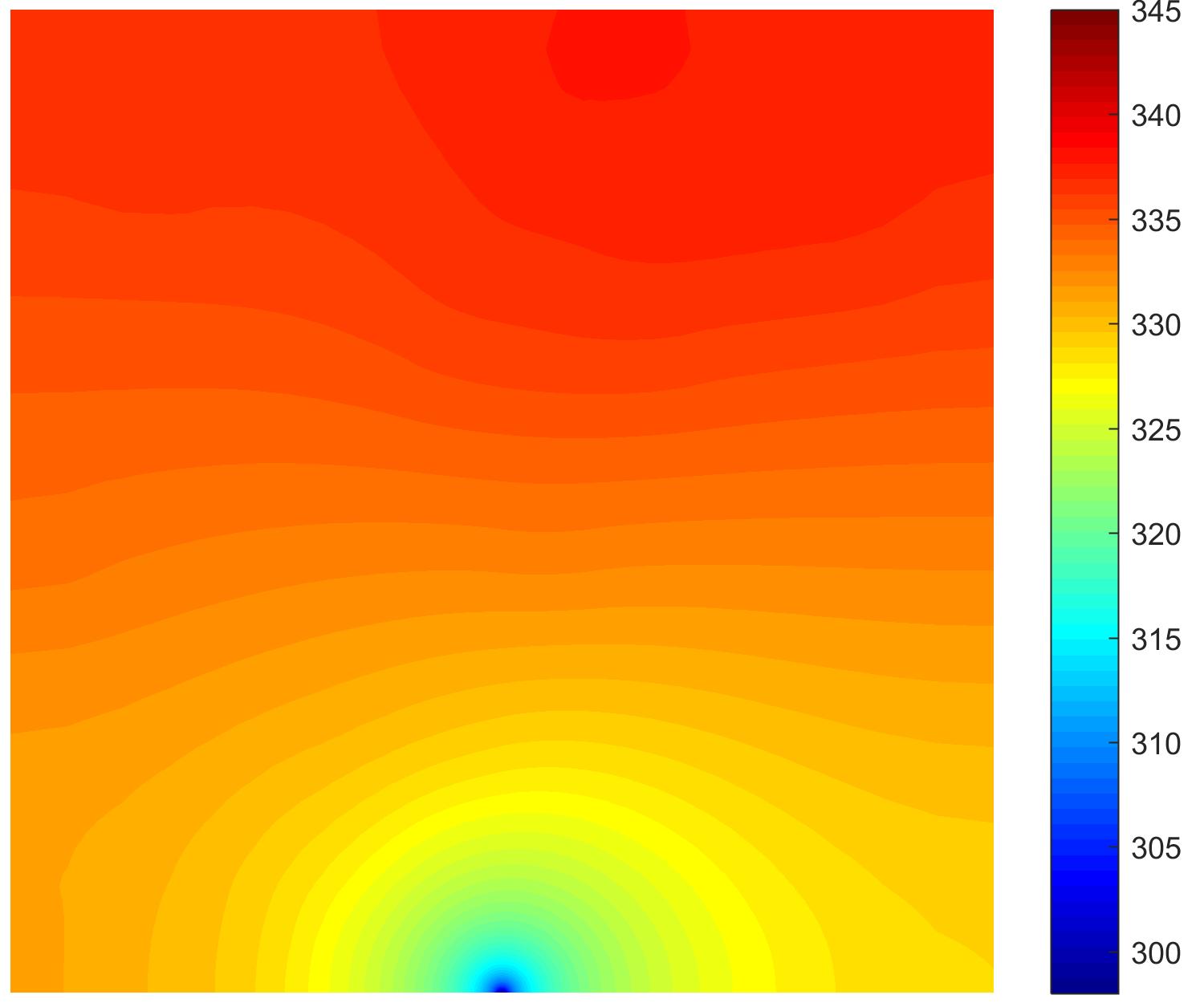}\label{fig9b}}
	\hspace{3mm}
	\subfigure
	{\includegraphics[scale=0.05]{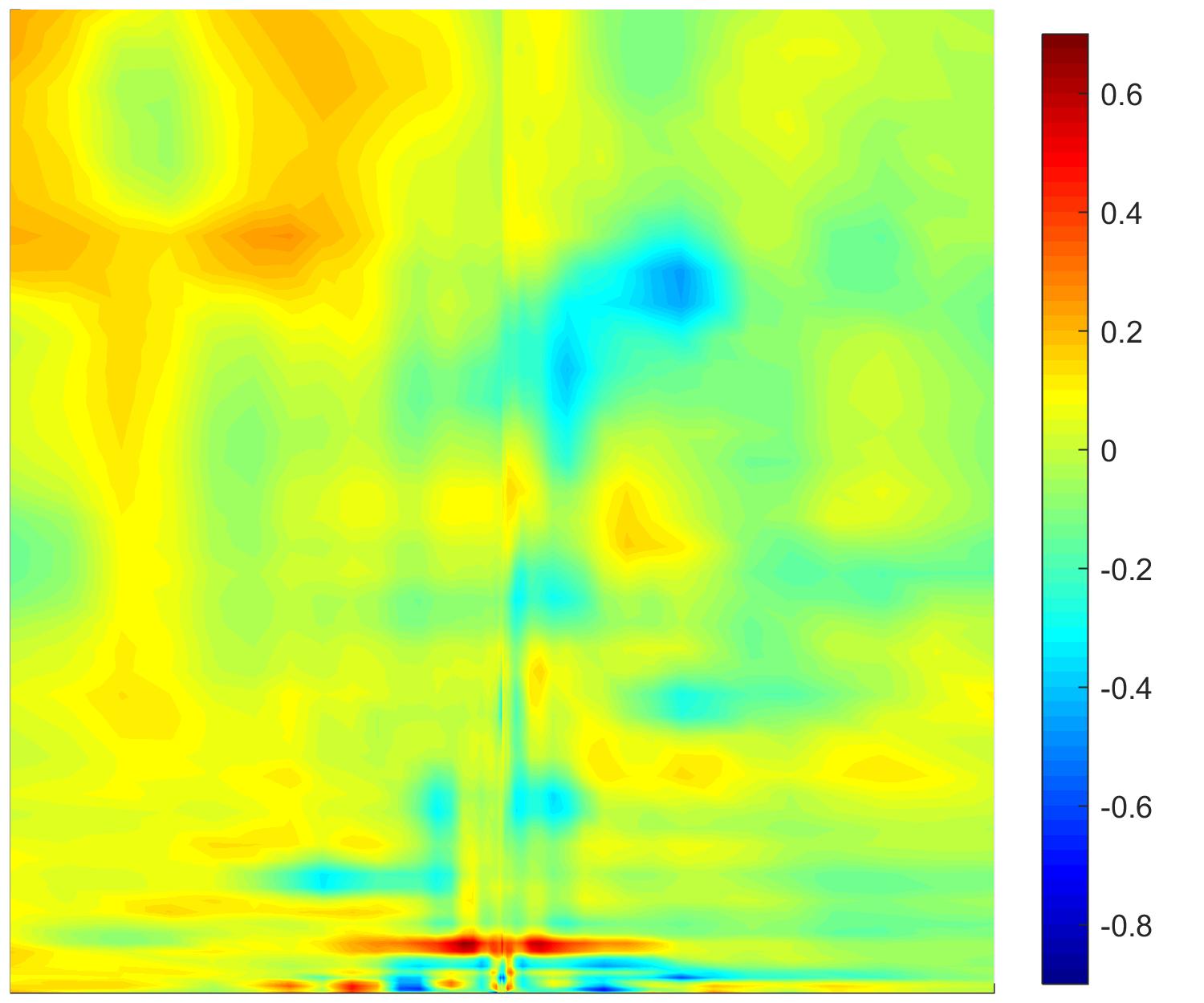}\label{fig9b}}
	
	\subfigure [Input layout]
	{\includegraphics[scale=0.05]{Definitions/mixpath1.jpg}\label{fig9a}} 
	\hspace{3mm}
	\subfigure [Predicted temperature] 
	{\includegraphics[scale=0.05]{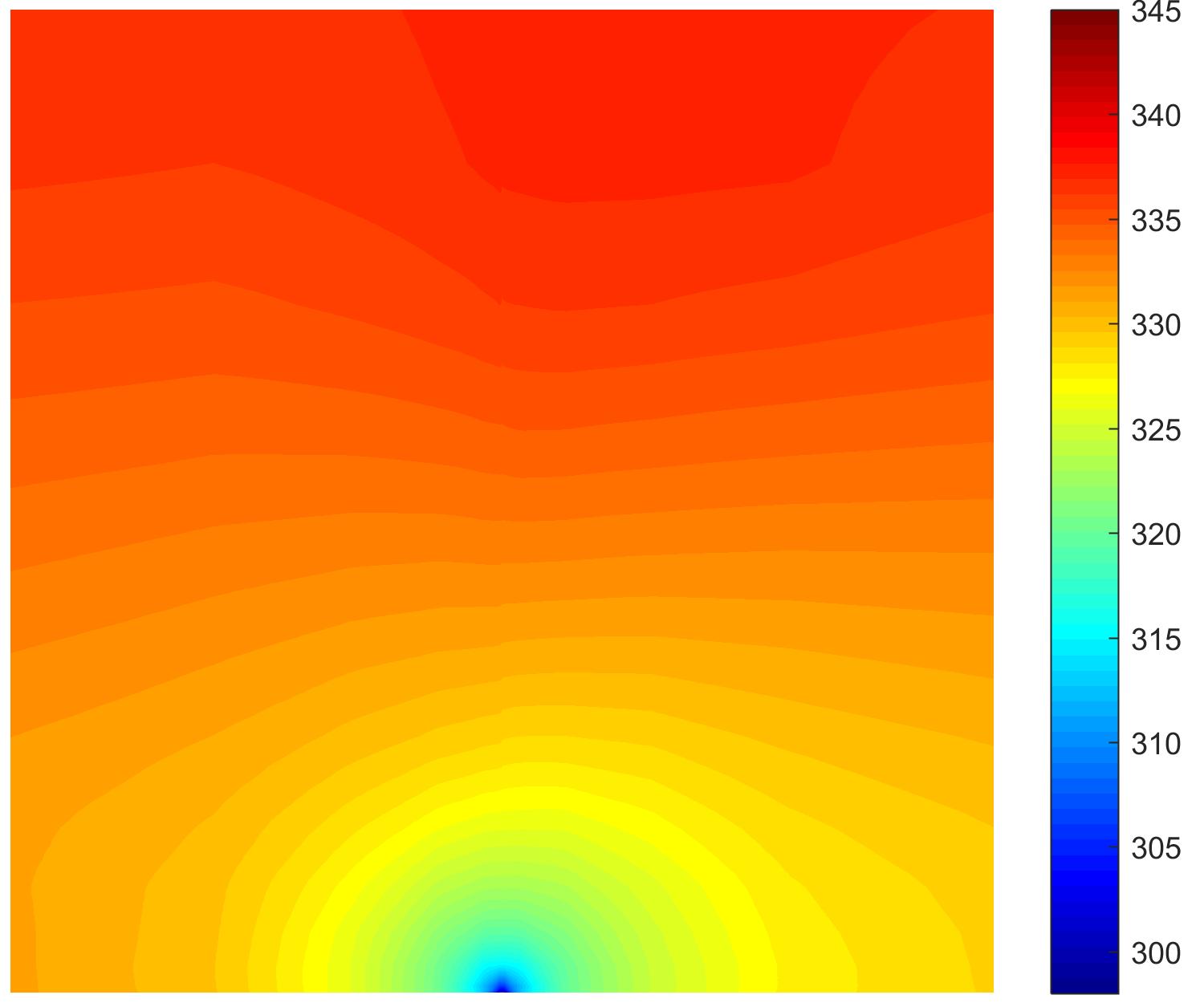}\label{fig9b}}
	\hspace{3mm}
	\subfigure [ Simulation]
	{\includegraphics[scale=0.05]{Definitions/mixpath4.jpg}\label{fig9b}}
	\hspace{3mm}
	\subfigure [ Error]
	{\includegraphics[scale=0.05]{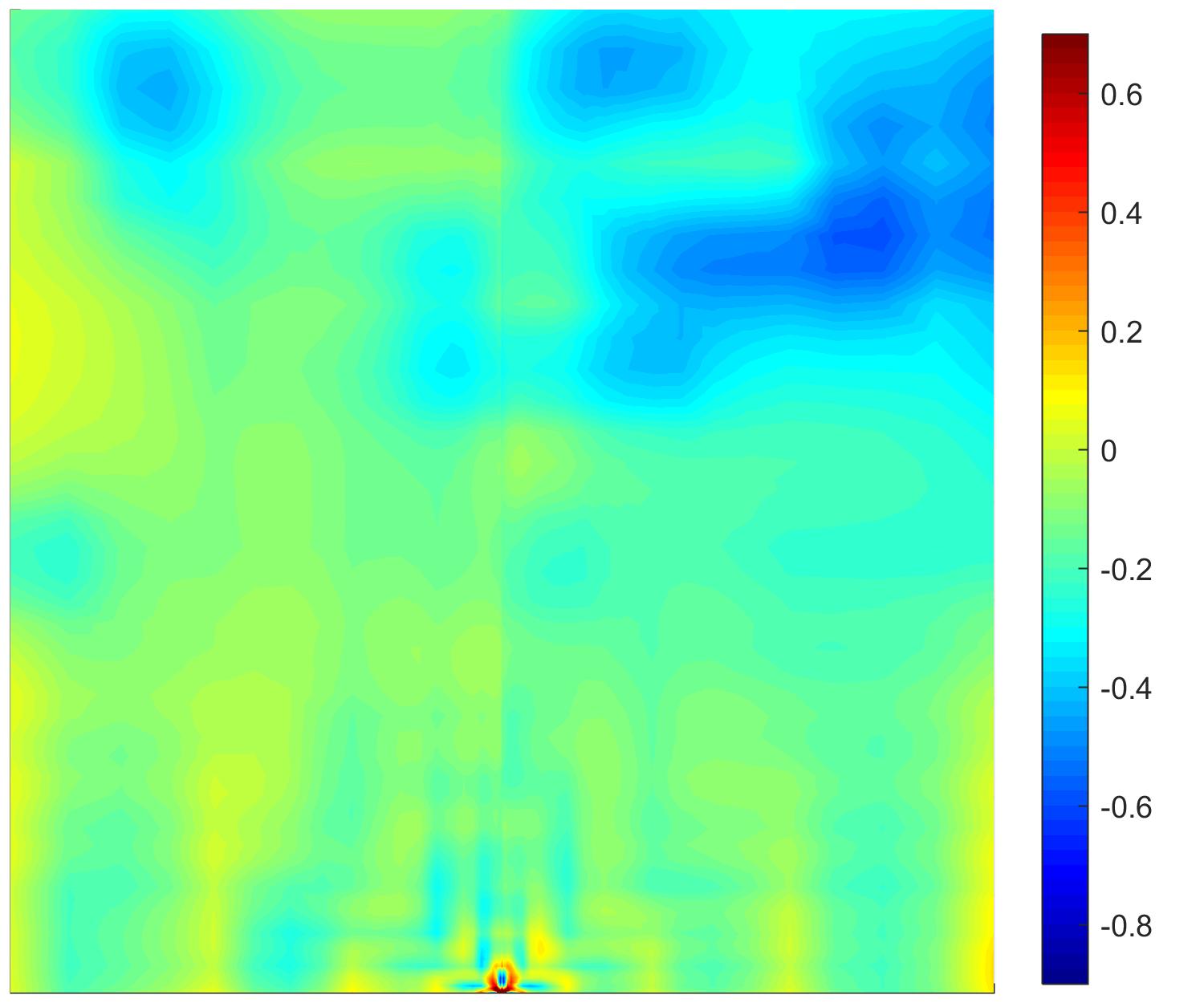}\label{fig9b}}
	\caption{The visualization of the prediction by using original FPN model (the first row) and searched Mixpath$\_$FPN model (the second row). (Temperature unit: K, the same below)}\label{fig8}
\end{figure*}

To illustrate the generality of the searched model compared with the original FPN, we randomly take one layout sample from the test set and let two models make the predictions. The visualizations of the input heat source layout, the predicted temperature field, the heat simulation of the corresponding layout and the error of between them are presented in Figure~\ref{fig8}. The corrsponding MAE and maximum AE are presented in Table~\ref{data}. From Figure~\ref{fig8} and Table~\ref{data}, it could be seen that the Mixpath$\_$FPN model possesses smaller MAE and maximum AE, which could be reduced from 0.1754K to 0.0998K, 1.82K to 0.86K respectively compared with the original FPN.  

\begin{table*}[h]
	\centering
	\caption{The MAE and maximum AE predicted by the original FPN model and the searched Mixpath$\_$FPN model on one layout sample.}
	
	\begin{tabular}{ccc}
		\toprule
		{Model} & {Mean Absolute Error} & Maximum Absolute Error \\ \midrule
		{\sc \ FPN} & 0.1754K   & 1.82K  \\
		{\ Mixpath$\_$FPN} & \textbf{0.0998K}   & \textbf{0.86K}  \\
		\bottomrule
	\end{tabular}
	
	\label{data}
\end{table*}

\subsubsection{MHSLO based on  the searched model}

After obtaining a deep learning surrogate model with less inference time and higher accuracy, then we use our designed MNSLO to identify the optimzal heat source layout scheme in case 1. To verify the effectiveness and improvement of the proposed MNSLO, we design two experiments. In the first experiment, we use the MNSLO based on the searched Mixpath$\_$FPN and the original FPN model respectively to test the improvement of total real optimiation time. Then in the second experiment, to test the performance of multimodal optimization, we make a comparison with  neighborhood search-based layout optimization (NSLO). NSLO is designed by Chen et al. \cite{Chen2020} to solve the same case.  The code of NSLO has been released \footnote{https://github.com/
	idrl-lab/HSLO-DLSM.}.
\begin{figure}[h]
	\centering
	\includegraphics[scale=0.06]{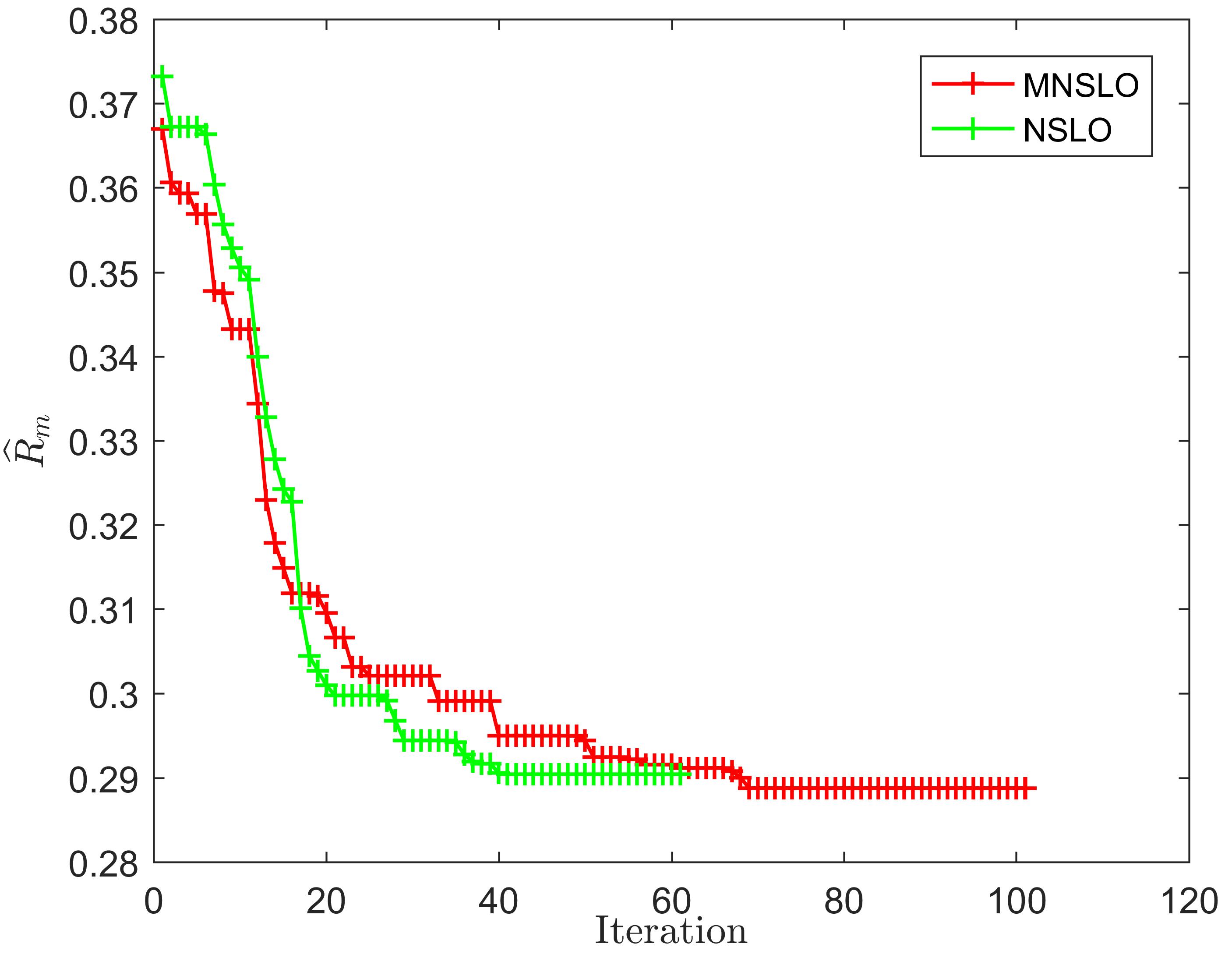}
	\caption{The iteration history of the NSLO algorithm and MNSLO algorithm for solving the case 1 based on Mixpath$\_$FPN surrogate.}
	\label{Iteration_NS1}
\end{figure}

To farily compare the effect of NSLO and MNSLO, the deep learning surrogates are both selected as the same in two experiments. To compare the global optimziation ability, we set the number of groups $c$ in MNSLO as 1, the convergence curves of two algorithms are presented in Figure \ref{Iteration_NS1}. As we can see, MNSLO finds better solution than NSLO, which is trapped in local optimum after 40 iterations. Then the real max temperature simulated by FDM is shown in Table \ref{Tab4}. The maximum temperature of founded heat source layout is optimized from 327.02K in NSLO to 326.74K. We also list the time cost of our method in Table \ref{traintime}. It shoud be noted that the total time including the data preparation and training the neural network is one-time cost. This means that once the deep learning surrogate is obtained after training, the inference capability can be used for good, which enables great exibility for real-time analysis. Besides, the less the inference time of neural network is, the larger the computational cost of whole optimization decreases.
\begin{table*}[h]
	\centering
	\caption{The statistic of time cost of our method.(time unit: hour)}
	\setlength{\tabcolsep}{1.5mm}{
		\begin{tabular}{cccc}
			\toprule
			
			{Process} & {Data preparation} & {Traing the supernet} & {Training the searched model}\\ \midrule
			{Time} & 2.905h   & 6.514h & 5.285h    \\
			\bottomrule
	\end{tabular}}
	
	\label{traintime}
\end{table*}

\begin{table*}[t]
	\scriptsize
	
	\centering
	
	\caption{The solutions of two heat source layout optimization design cases solved by MHSLO-NAS (Ours) algorithm and compared with the presented best solutions in literature.}
	
	\label{Tab4}
	\setlength{\tabcolsep}{1.5mm}{
		\begin{tabular}{cccccccccc}
			
			\toprule
			
			\multirow{2}{*}{$Case$}&	\multirow{2}{*}{$Method$} & \multicolumn{2}{c}{$Simulation$} & \multicolumn{2}{c}{$Prediction$} & \multicolumn{2}{c}{$Error$} \\
			
			\cmidrule(r){3-4} \cmidrule(r){5-6} \cmidrule(r){7-8}
			
			&	&  $T_{\max }$(K) 	&  $R_{m}$   &   $\widehat{T}_{\max }$(K) 		
			&  $\widehat{R}_{m }$      &  $\Delta \widehat{T}_{\max }$(K)    &   MAE(K)   \\		
			\midrule		
			\multirow{8}{*}{Case 1}&	{MHSLO-NAS(1)}   &\textbf{326.83}   & 0.2883    & 326.85     & 0.2885            & 0.02     &0.1034           \\	
			&	{MHSLO-NAS(2)}     &\textbf{326.94}         & 0.2894     & 326.86      & 0.2886   & 0.08        & 0.1140          \\	
			&	{MHSLO-NAS(3)}      &\textbf{326.74}                          & 0.2874        & 326.87                   & 0.2887           & 0.13                  & 0.1147           \\	
			&	{MHSLO-NAS(4)}              &\textbf{326.95}                          & 0.2895                    & 326.99                 & 0.2899          & 0.04             & 0.1155            \\
			\cline{2-8}
			&	\text { Aslan et al. }\cite{Aslan}   &328.05      & 0.3005  & -  & - & -  & - \\
			&	\text { Chen et al. } \cite{Chen} &328.69 & 0.3069   & -    & -         & -  & -           \\
			&	\text { Chen et al. }\cite{Chen2020}   &327.02   & 0.2902    & 327.04     & 0.2904            & 0.02     &0.1294             \\
			\midrule
			\multirow{8}{*}{Case 2}&	{MHSLO-NAS(5)}   &\textbf{328.89}   & 0.3089    & 328.04      & 0.3004           & 0.85     & 0.2133          \\	
			&	{MHSLO-NAS(6)}     &\textbf{328.89}         & 0.3089     & 328.03      & 0.3003  & 0.86        & 0.1210          \\	
			&	{MHSLO-NAS(7)}      &\textbf{328.95}                        & 0.3095        & 328.02                   & 0.3002          & 0.92                   & 0.1817           \\	
			&	{MHSLO-NAS(8)}              &\textbf{328.97}                              & 0.3097                  & 328.02                   & 0.3002          & 0.95             & 0.1955            \\
			\cline{2-8}
			&	\text { Chen et al. }\cite{Chen2020}  &333.51      & 0.3351  & 333.12 & 0.3312 & 0.39  & 0.1934 \\
			\bottomrule		
	\end{tabular}}
\end{table*}

\begin{figure}[H]
	\centering
	\includegraphics[scale=0.9]{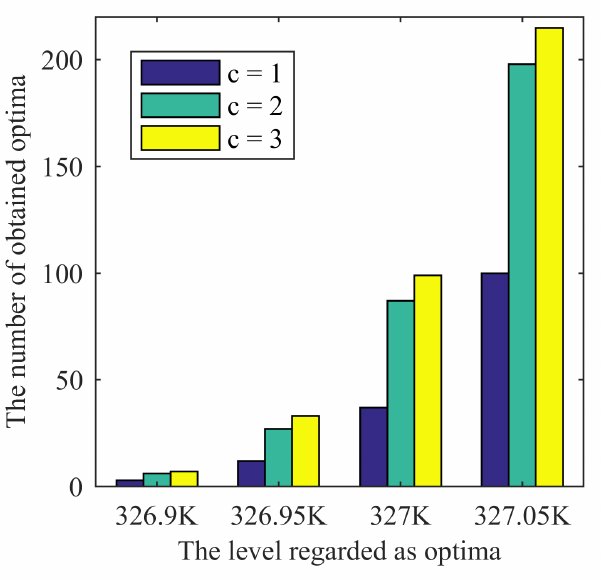}
	\caption{The number of the obtained layout design schemes in case 1 by MNSLO.}
	\label{multimodal}
\end{figure}

\begin{figure}[h]
	\centering
	\subfigure [Layout]
	{\includegraphics[scale=0.05]{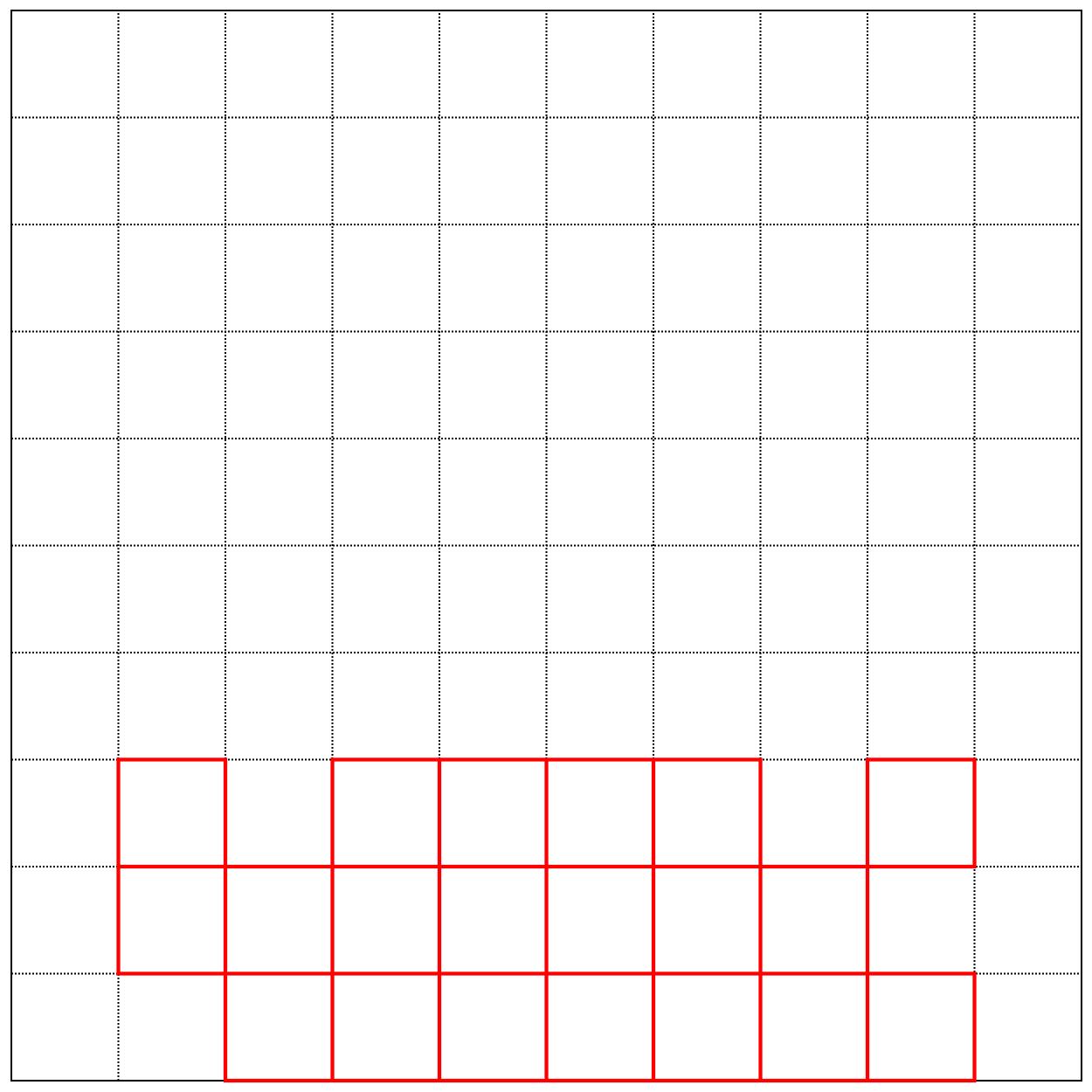}\label{fig9a}} 
	\hspace{3mm}
	\subfigure [Simulation] 
	{\includegraphics[scale=0.05]{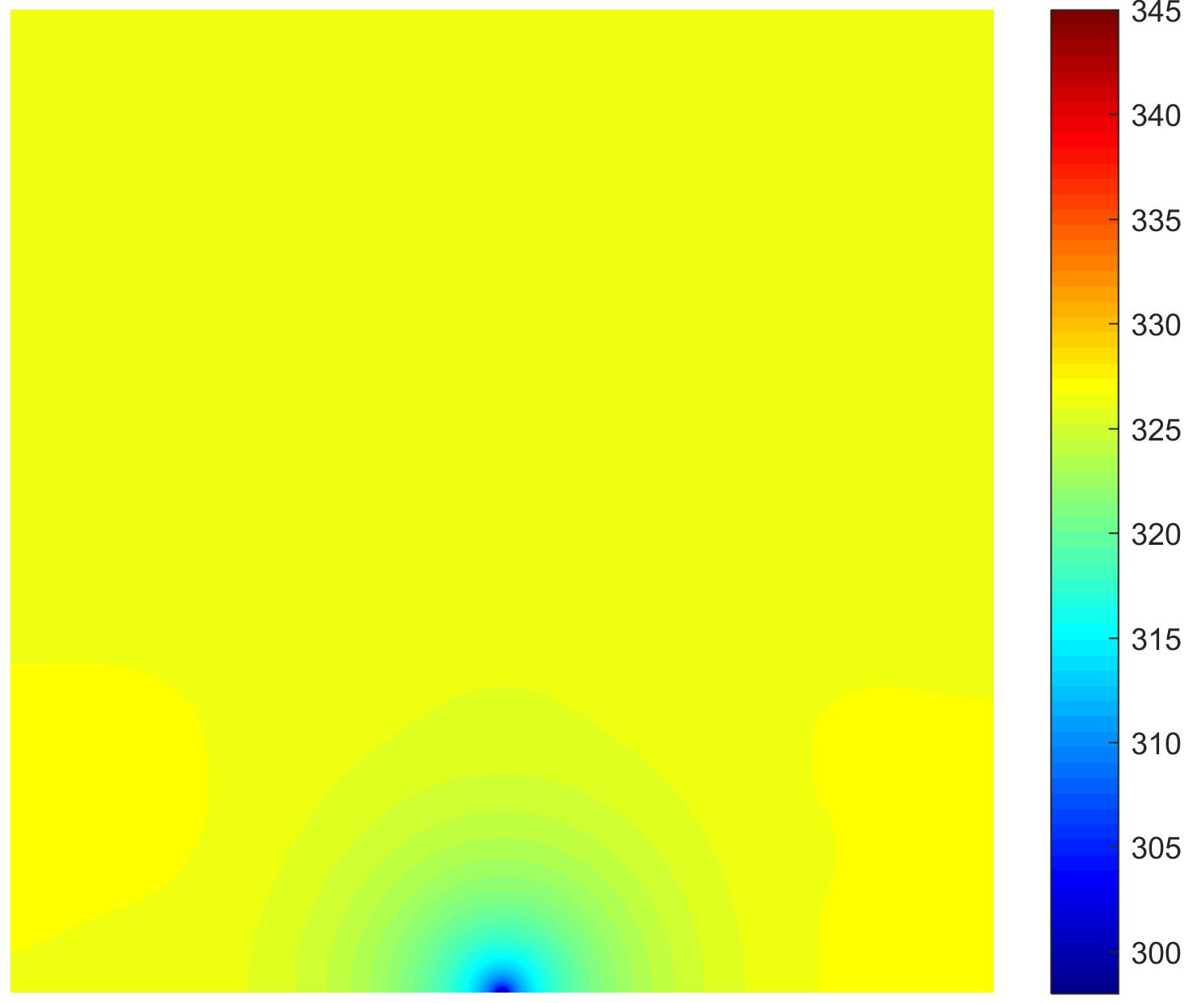}\label{fig9b}}
	\hspace{3mm}
	\subfigure [ Layout]
	{\includegraphics[scale=0.05]{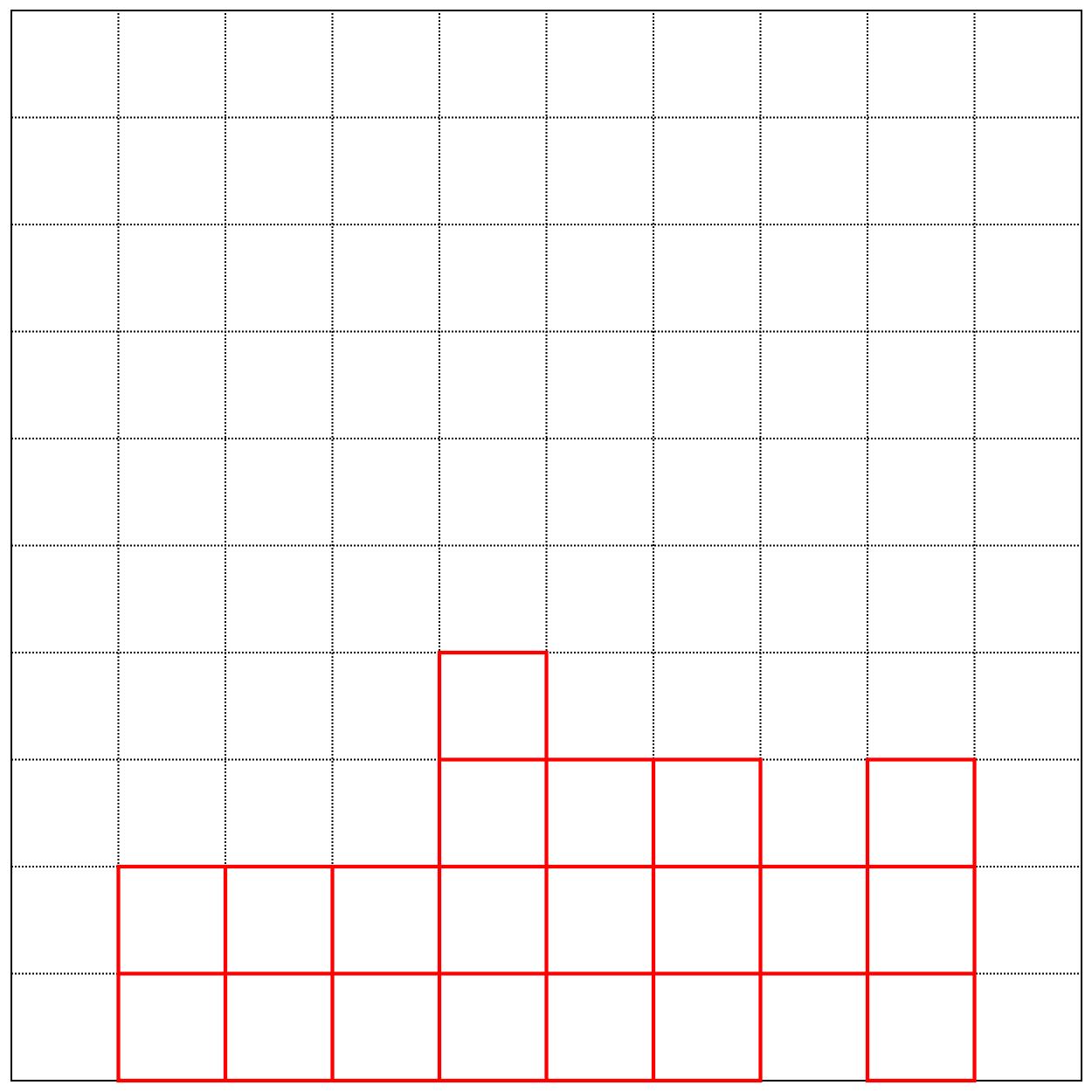}\label{fig9b}}
	\hspace{3mm}
	\subfigure [ Simulation]
	{\includegraphics[scale=0.05]{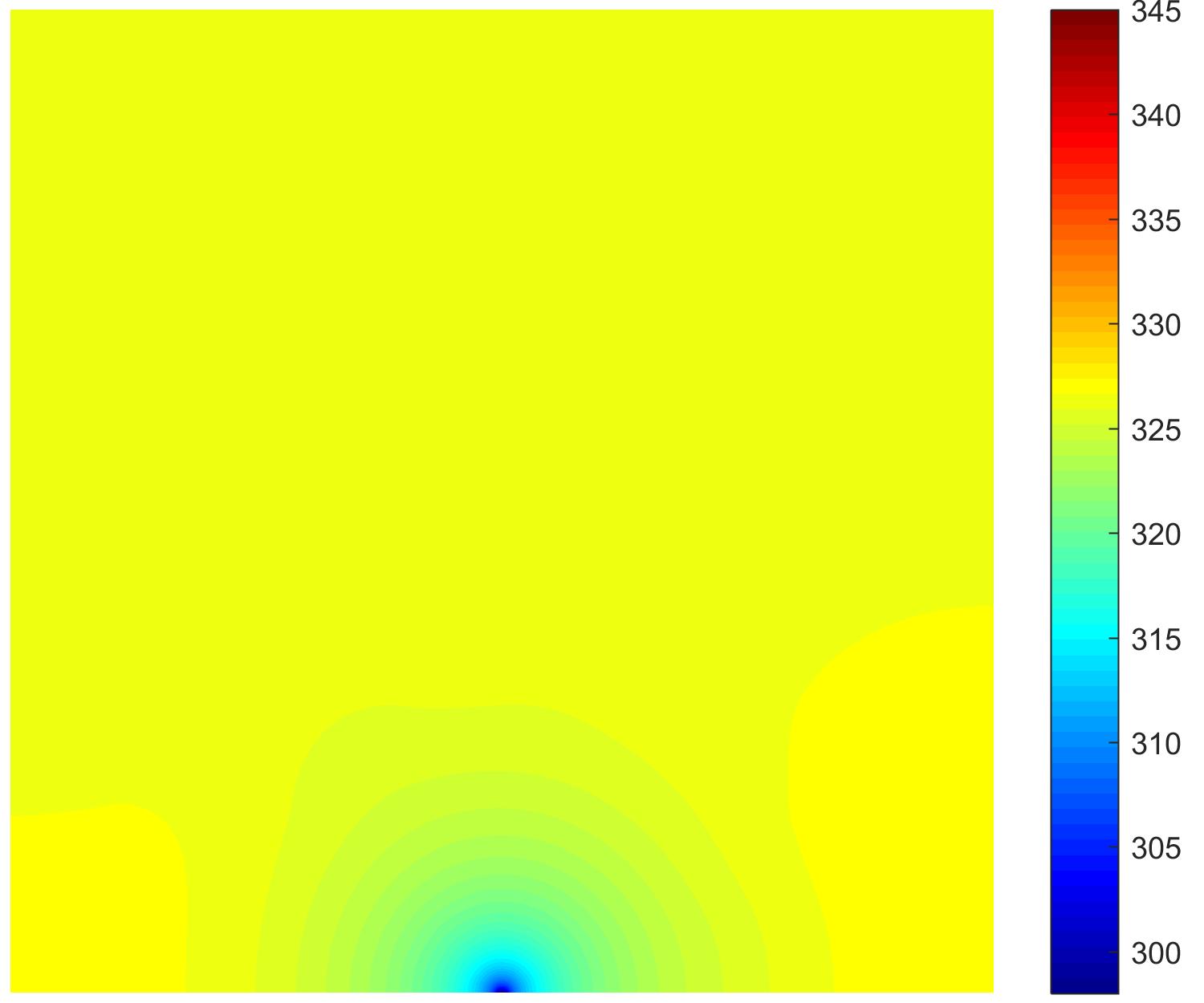}\label{fig9b}}
	
	\subfigure [Layout]
	{\includegraphics[scale=0.05]{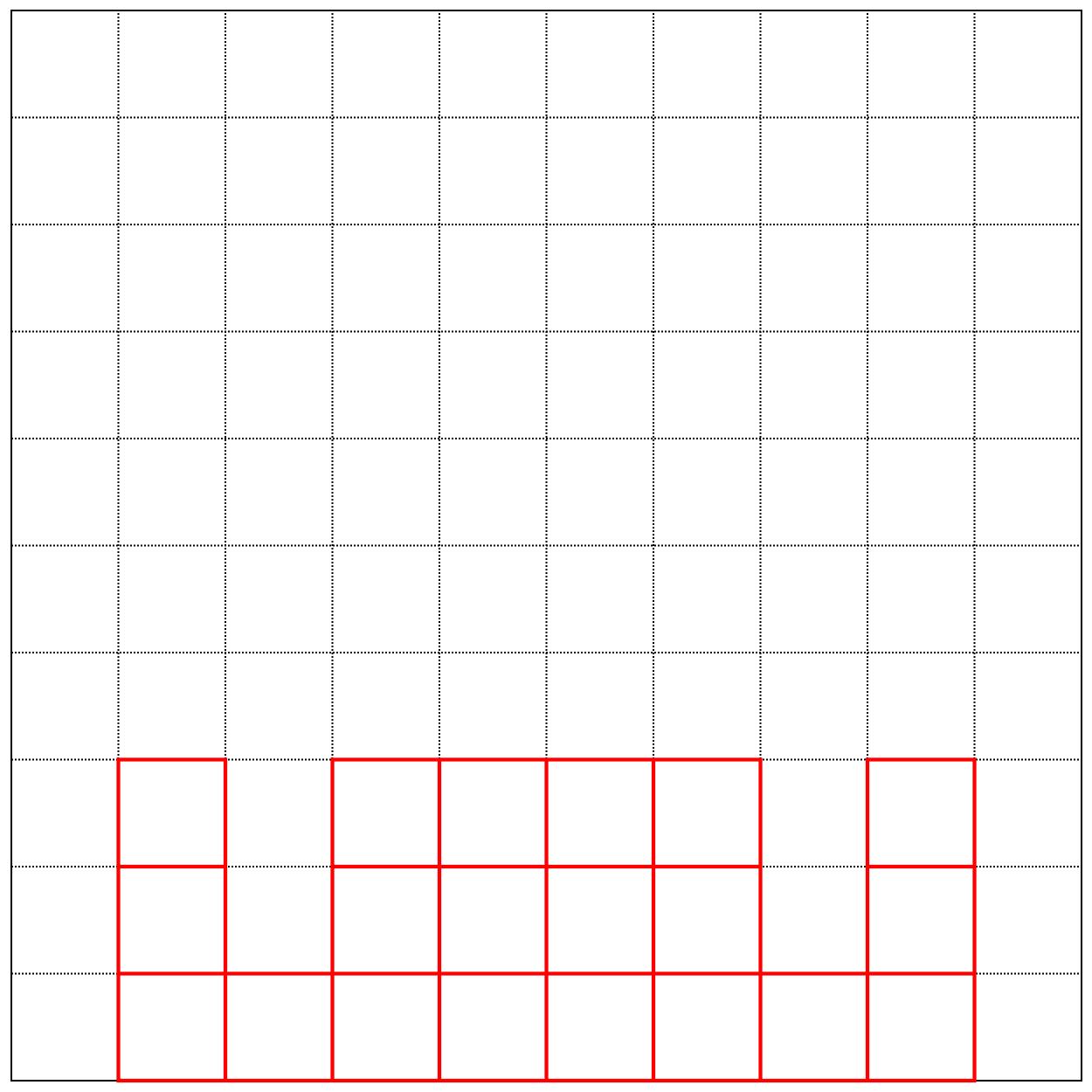}\label{fig9a}} 
	\hspace{3mm}
	\subfigure [Simulation] 
	{\includegraphics[scale=0.05]{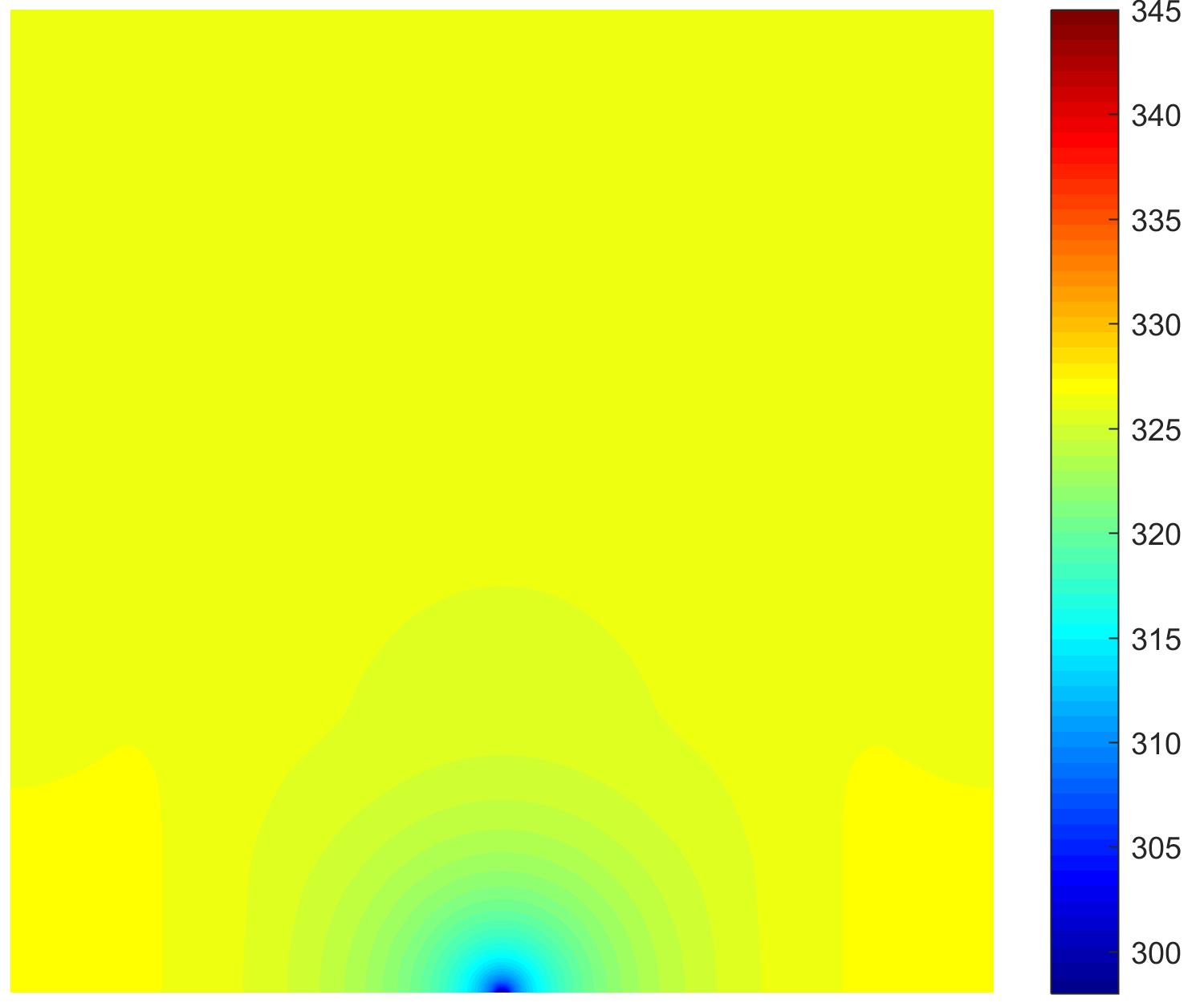}\label{fig9b}}
	\hspace{3mm}
	\subfigure [ Layout]
	{\includegraphics[scale=0.057]{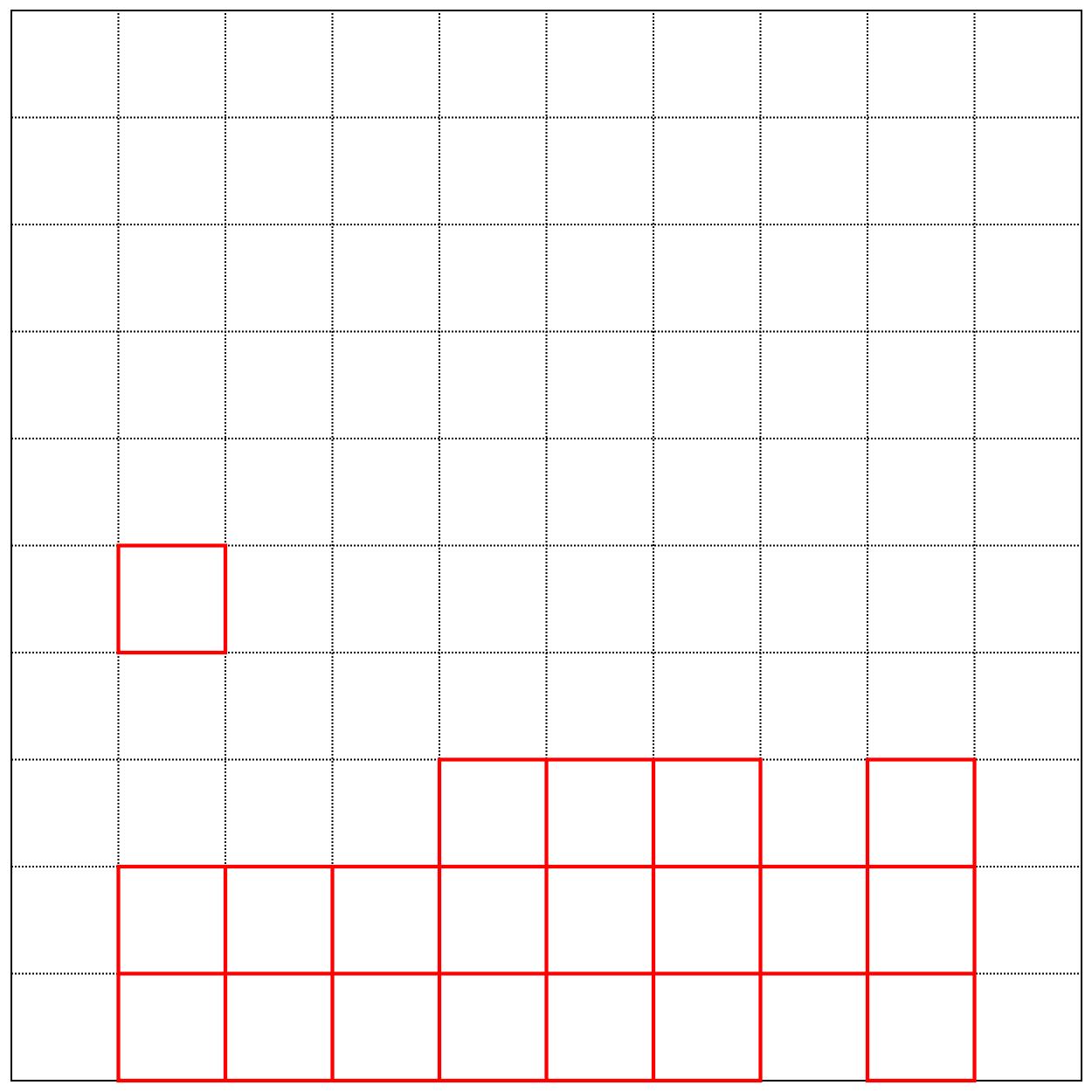}\label{fig9b}}
	\hspace{3mm}
	\subfigure [ Simulation]
	{\includegraphics[scale=0.05]{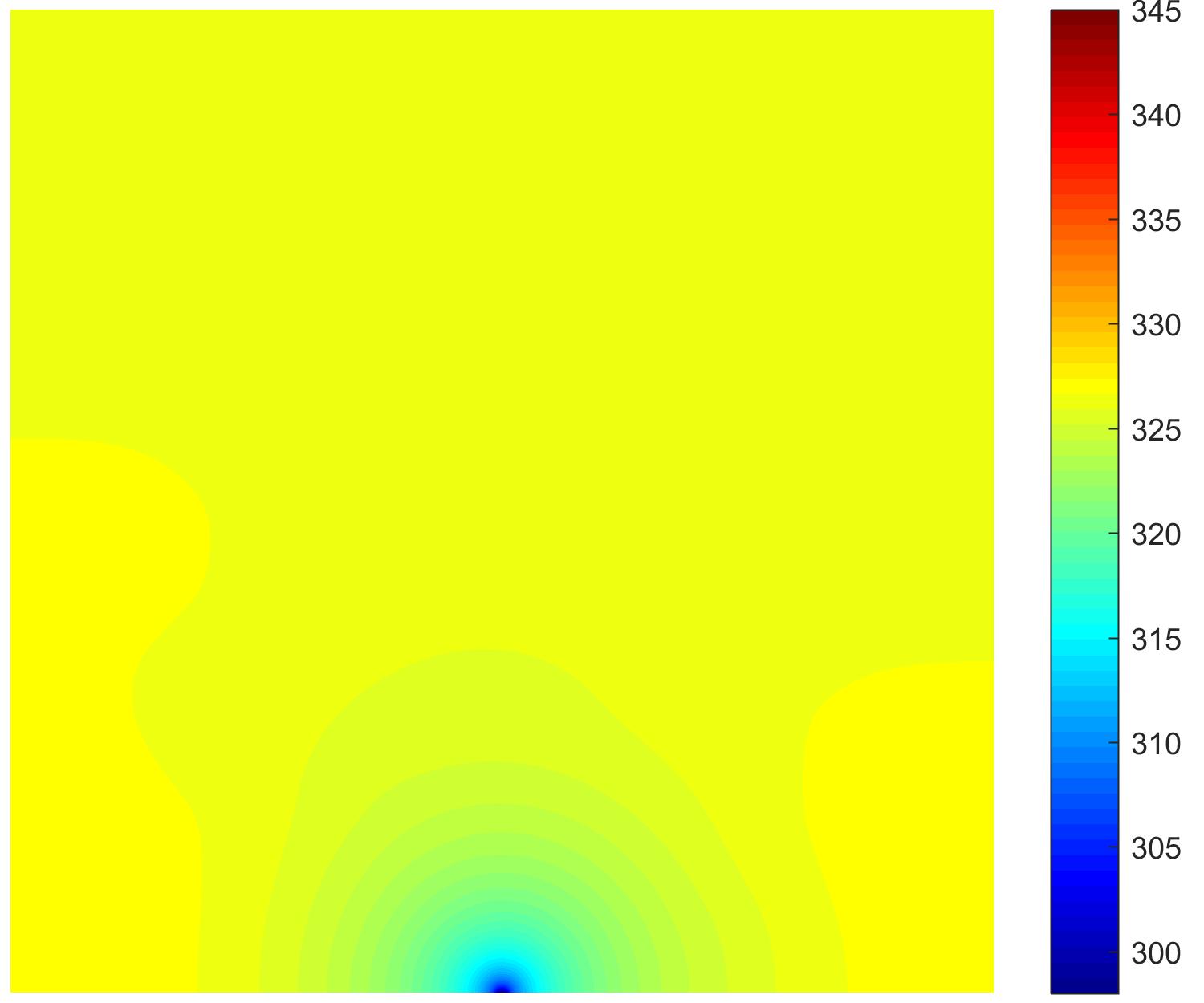}\label{fig9b}}
	\caption{An illustration of  four heat source layout schemes obtained by our method and their corresponding simulation in  case 1. Left top is MHSLO-NAS(1), Right top is MHSLO-NAS(2), Left bottom is MHSLO-NAS(3), Right bottom is MHSLO-NAS(4) }\label{fig9}
\end{figure}

Apart from obtaining the better solution compared with previous work, we further test the multimodal optimization effect of our proposed method. We set a threshld value of the maximum temperature as the level regarded as the optimal solutions. Then we evaluate the performance from the number of optimal solutions obtained by us. We set the threshold to 326.9K, 326.95K, 327K and 327.05K respectively, which are all lower than the result reported in Chen et al.\cite{Chen2020}. We also set the number of groups to 1, 2 and 3 to make a simple comparison. The result is presented in Figure \ref{multimodal}. As we can see, even though the threshold is set to 326.9K, we still could seek 3, 6 and 7 candidate solutions respectively, which is farther lower than 327.04K reported in Chen et al.\cite{Chen2020}. In addtion, when the number of groups is set to be larger, we could find more candidate optimal solutions. As the number of groups $c$ gets larger, the increasing of the number of obtained near optimal solutions slows down. Figure \ref{fig9} shows the four heat source layout schemes searched by our proposed method. 


\subsection{Case 2: heat source layout optimization with different heat intensity}

To further verify the effectiveness of our proposed method, we also define the HSLO problem with different intensities. The parameters of 20 heat source are presented in Table \ref{case2}. The heat intensity ranges from 2000W/${m}^{2}$ to 20000W/${m}^{2}$. Every two heat source share a kind of heat intensity.

\begin{table}[h]
	\scriptsize	
	\centering	
	\caption{The parameters of heat intensity of 20 heat sources in case 2 (intensity unit: W/${m}^{2}$) }
	
	\label{case2}

	\begin{tabular}{cccccccccc}	
		\toprule
		
		Intensity& 2000	&  4000 	&  6000   &  8000	
		&  10000      \\		
		\midrule		
		{number}   & 2         & 2           & 2    & 2    &2     \\
				\hline	
		{Intensity}         & 12000       & 14000   & 16000        & 18000   & 20000      \\ 
		\midrule		
		{number}   & 2         & 2           & 2    & 2     & 2    \\	
			
		\bottomrule		
	\end{tabular}	
\end{table}

\subsubsection{The performance of the searched model}

\begin{figure}[h]
	\centering
	\includegraphics[scale=1]{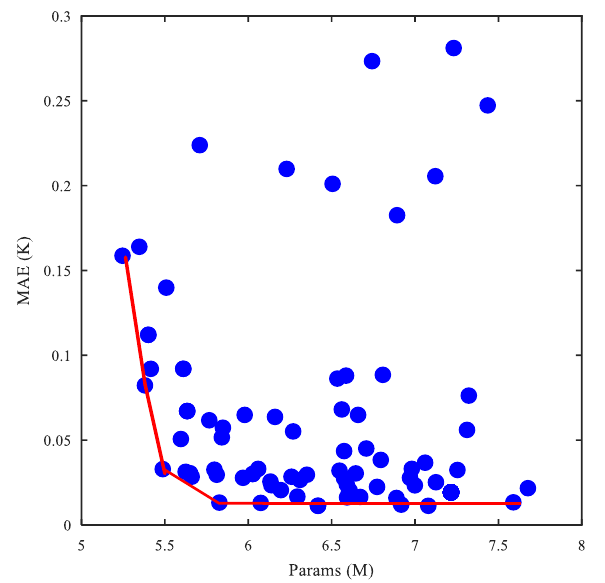}
	\caption{The pareto front of the searched models by NSGA-II in case 2.}
	\label{nsga2}
\end{figure}

\begin{figure}[h]
	\centering
	\includegraphics[scale=0.5]{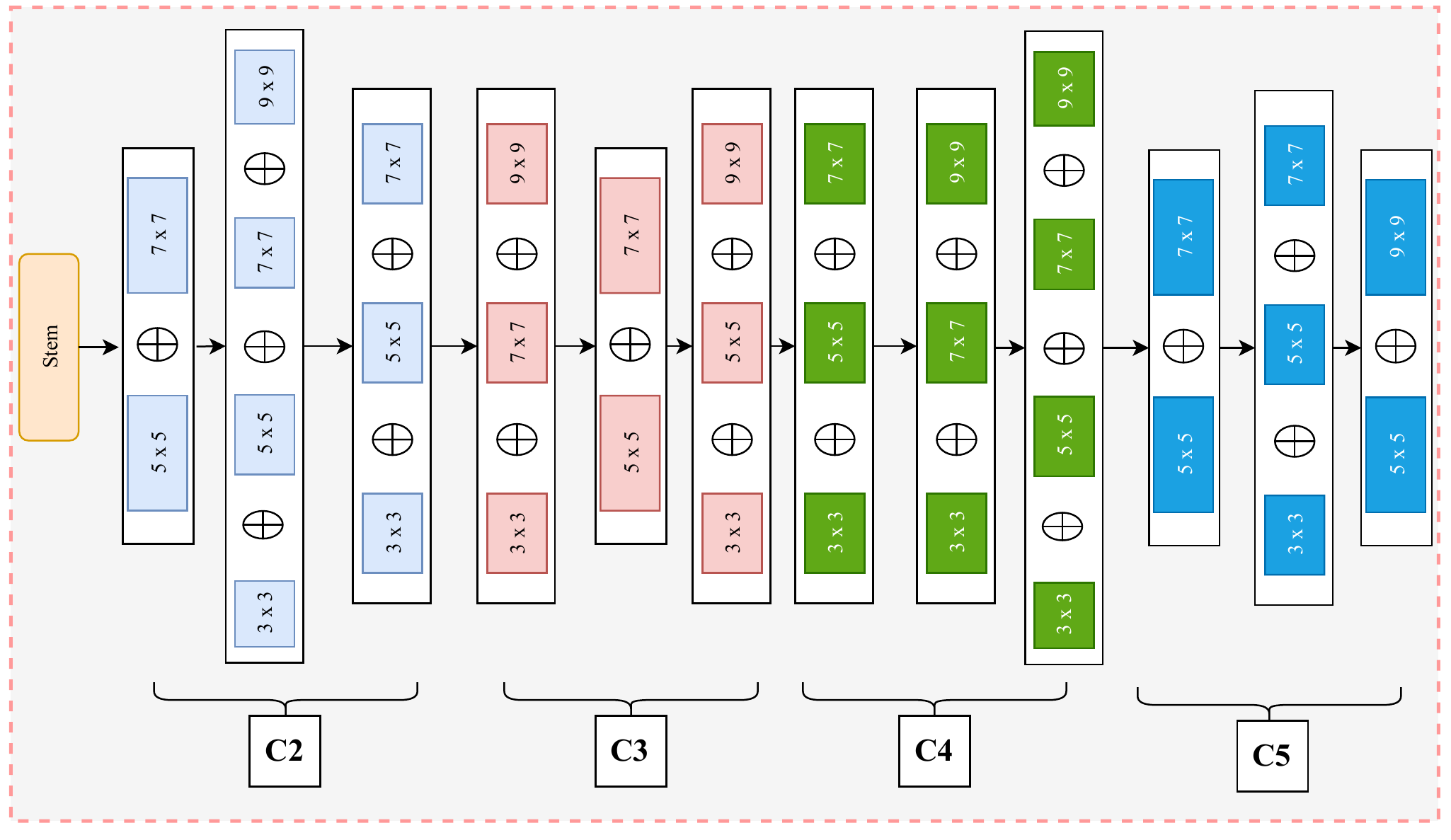}
	\caption{The visualization of the searched backbone architecture in case 2}
	\label{arch2}
\end{figure}

This defined case becomes more complex, thus it is more difficult to find the optimal layout scheme. However, in the process of training the surrogate, it does not brings out more challenges compared with case 1 due to purely regarding it as an image-to-image regression task. So in our experiments, the parameters are all set the same as case 1.  The result optimized by NSGA-II is presented in Figure \ref{nsga2}.  We also select one model architecture from the Pareto frontier. The searched architecture is presented in Figure \ref{arch2}.   

\begin{figure*}[h]
	\centering
	\subfigure 
	{\includegraphics[scale=0.05]{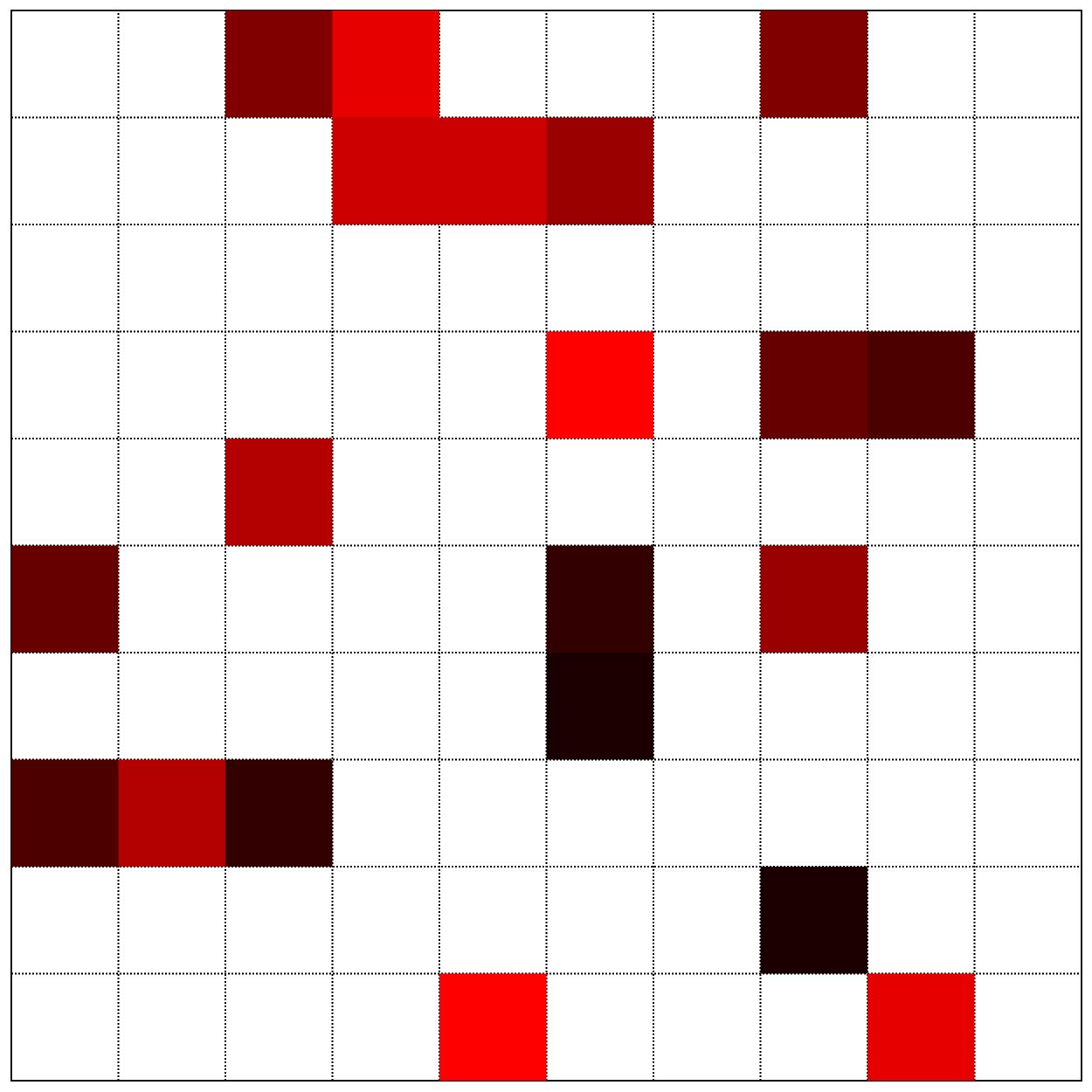}\label{fig9a}} 
	\hspace{3mm}
	\subfigure  
	{\includegraphics[scale=0.05]{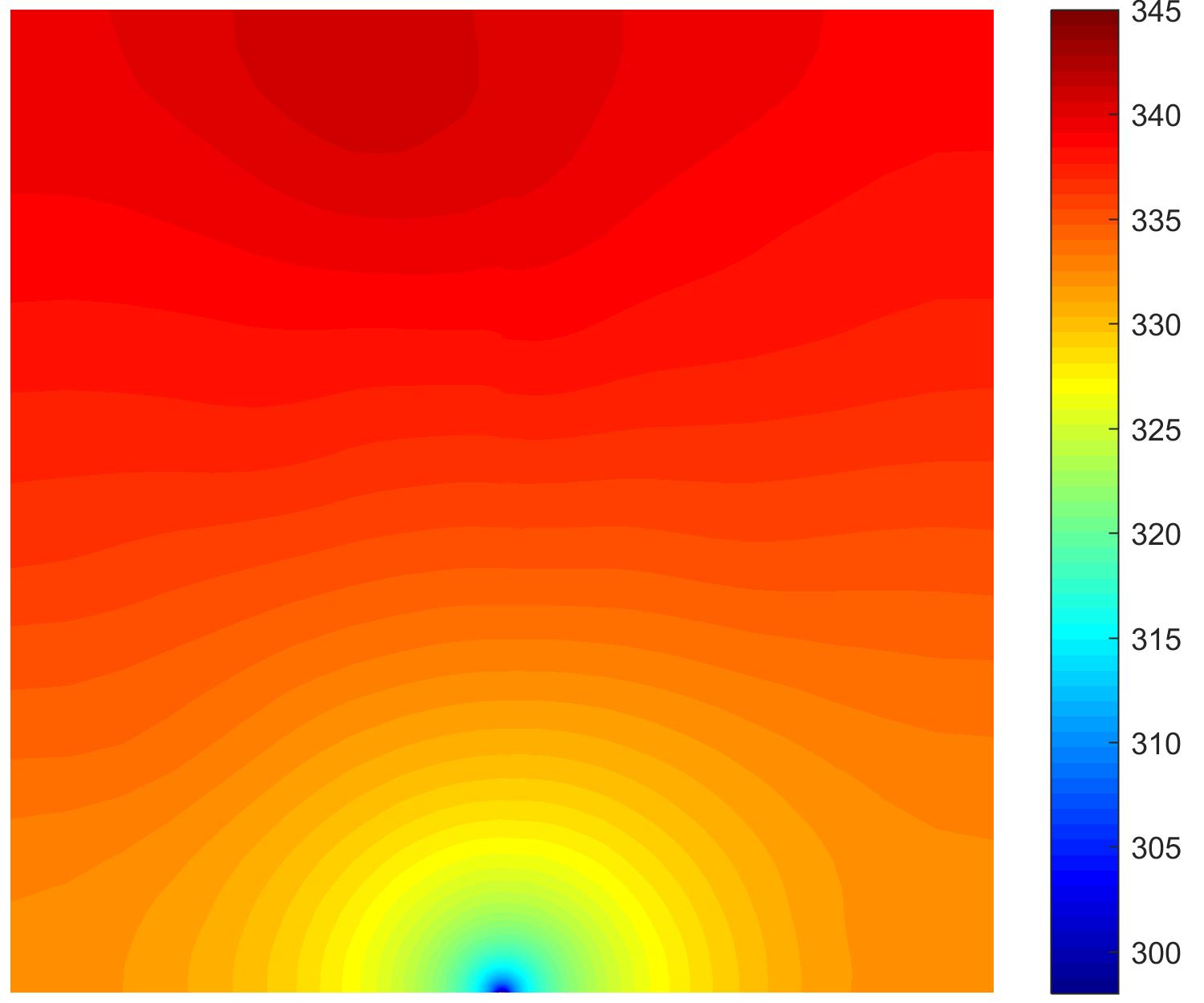}\label{fig9b}}
	\hspace{3mm}
	\subfigure 
	{\includegraphics[scale=0.05]{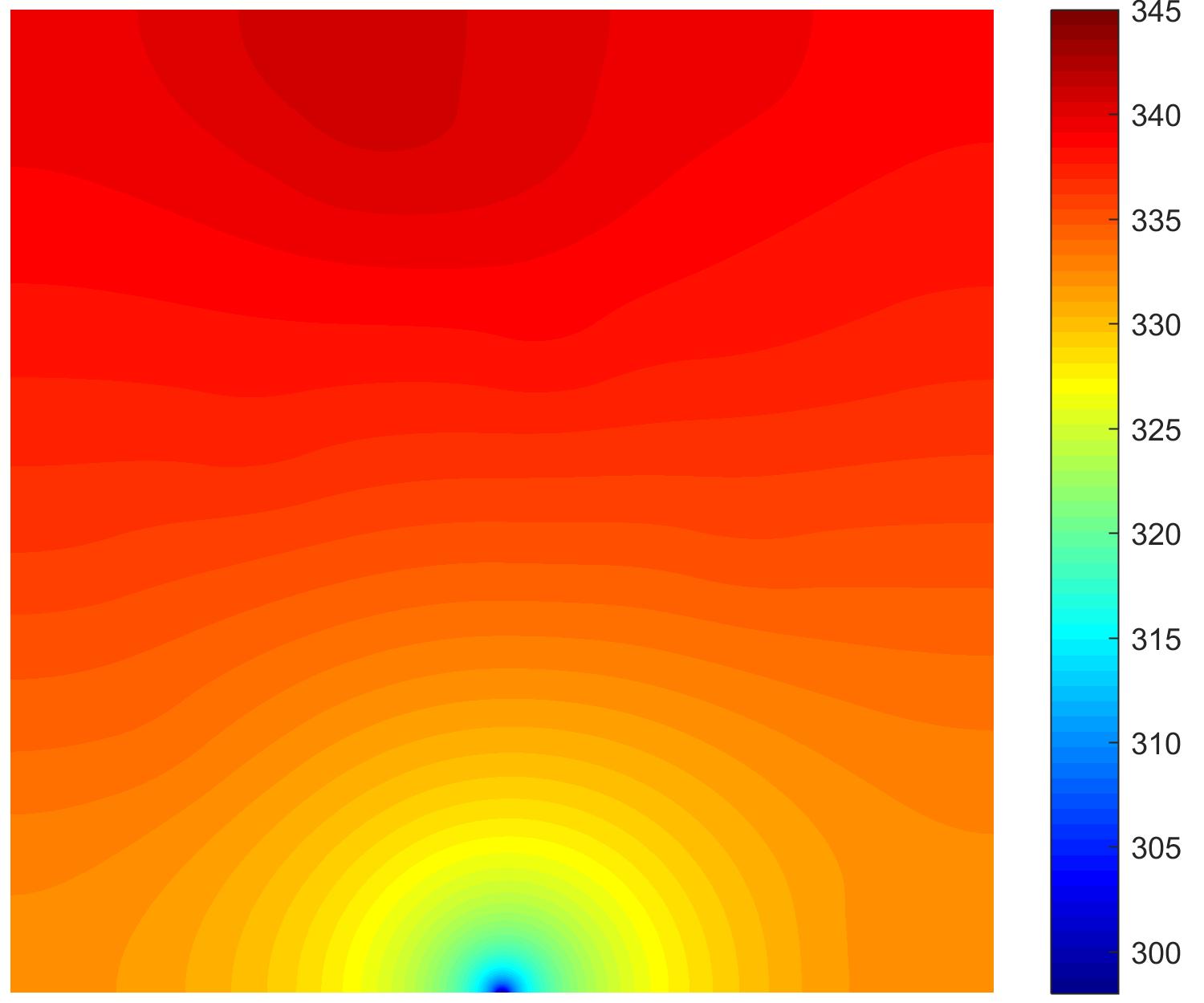}\label{fig9b}}
	\hspace{3mm}
	\subfigure
	{\includegraphics[scale=0.05]{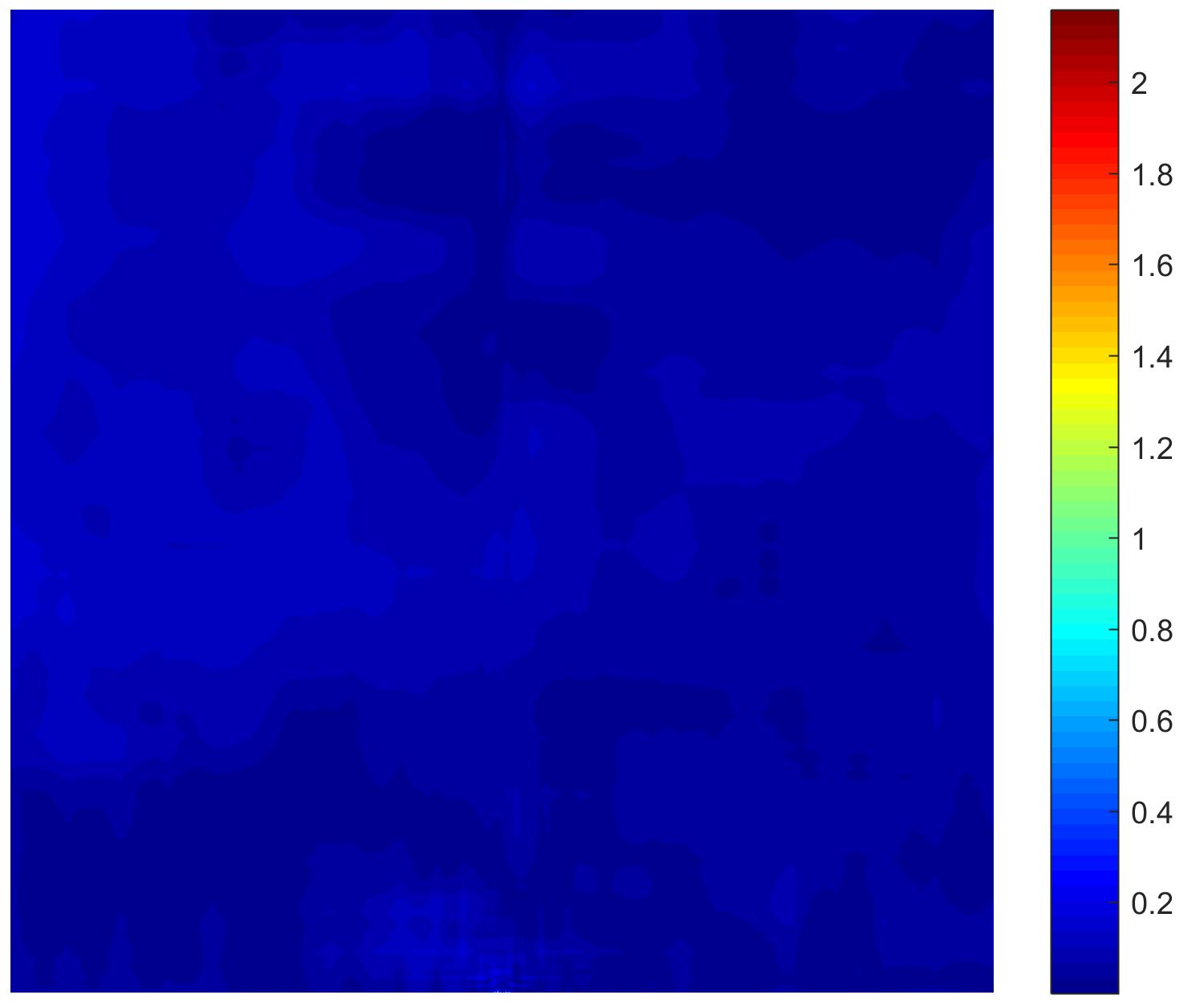}\label{fig9b}}
	
	\subfigure [Input layout]
	{\includegraphics[scale=0.05]{Definitions/kw_layout.jpg}\label{fig9a}} 
	\hspace{3mm}
	\subfigure [Predicted temperature] 
	{\includegraphics[scale=0.05]{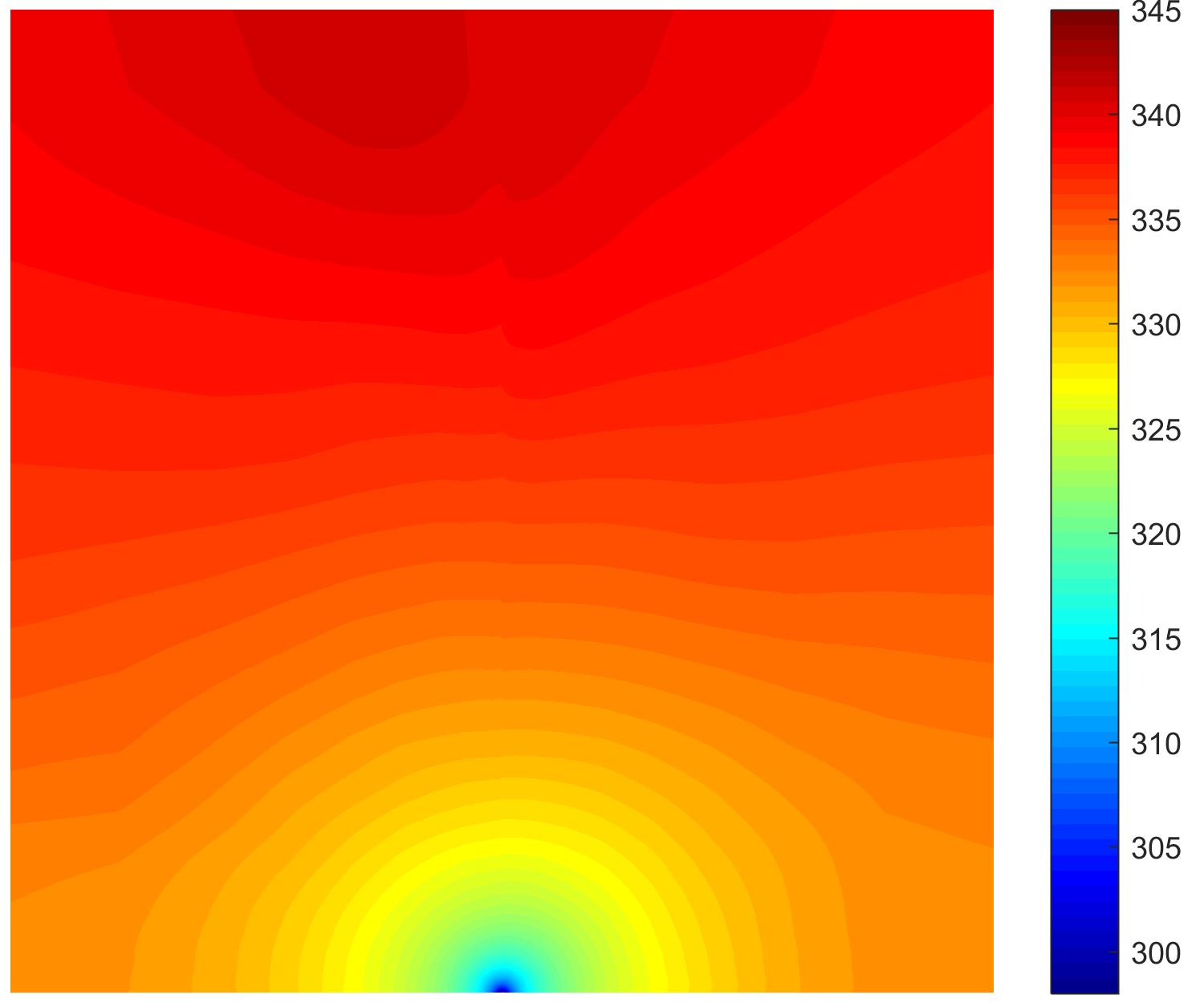}\label{fig9b}}
	\hspace{3mm}
	\subfigure [ Simulation]
	{\includegraphics[scale=0.05]{Definitions/kw_realheat.jpg}\label{fig9b}}
	\hspace{3mm}
	\subfigure [ Error]
	{\includegraphics[scale=0.05]{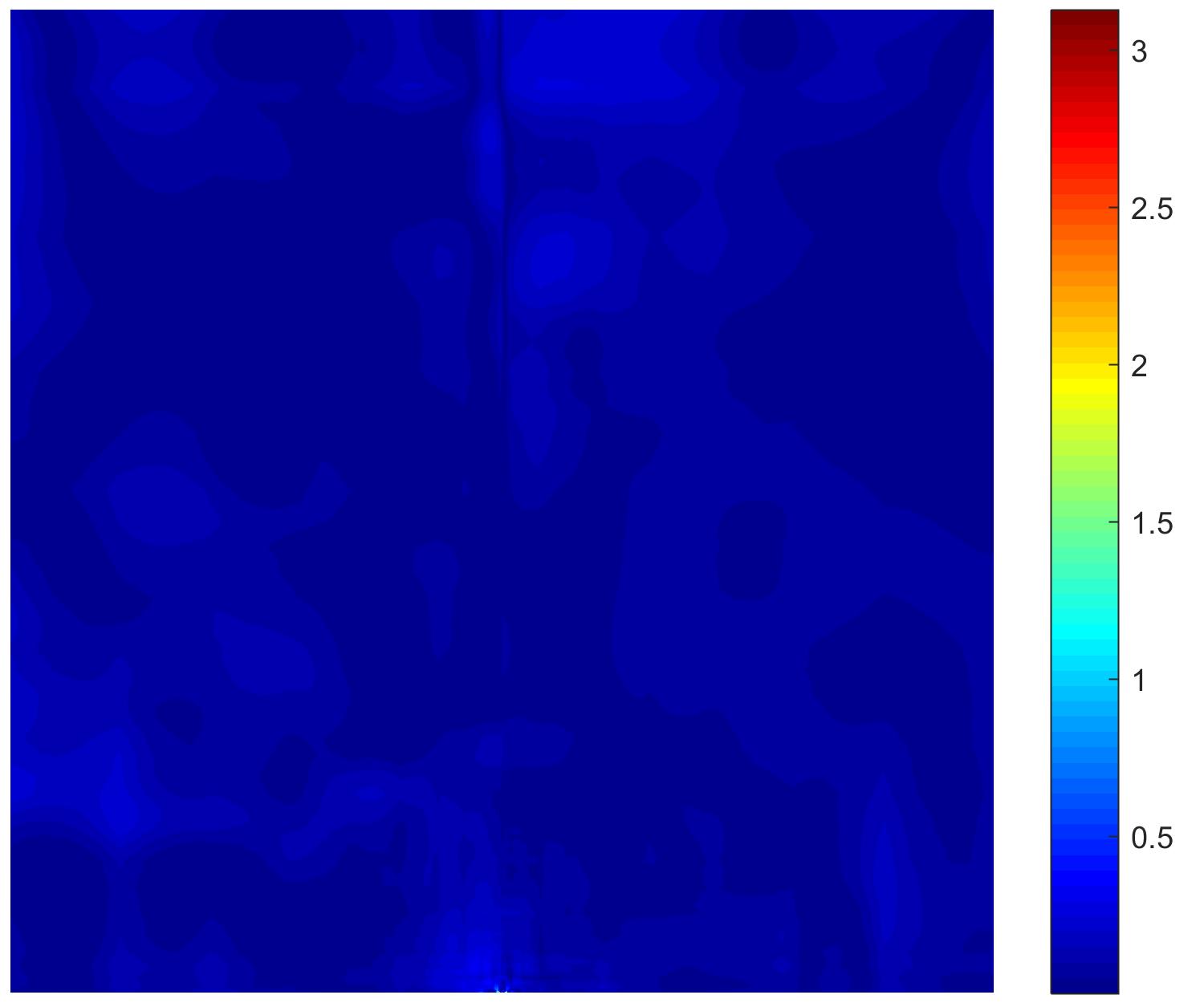}\label{fig9b}}
	\caption{The visualization of the prediction by using original FPN model (the first row) and searched Mixpath$\_$FPN model (the second row).}\label{fig99}
\end{figure*}

To evaluate the effect of NAS method, we test the four metrics introduced in Section~\ref{sec41} of the original FPN model and the searched Mixpath$\_$FPN model to make a comparison. The results are listed in Table~\ref{Tab03}. It could be seen that with similar predction accuracy, the total parameters size of the searched model is only 1/4 times the original FPN. To illustrate the generality of the searched model, we randomly take one layout sample from the test set and utilize the searched model and original FPN make predictions respectively. The visualization of the input heat source layout, the prediction, the heat simulation and the error of between them is presented in Figure~\ref{fig99}. From Figure~\ref{fig99}, it could be seen that Mixpath$\_$FPN model could learn the mapping from the layout to temperature field well, which possesses the similar accuracy with FPN.

\begin{figure*}[h]
	\centering
	\subfigure [Layout]
	{\includegraphics[scale=0.05]{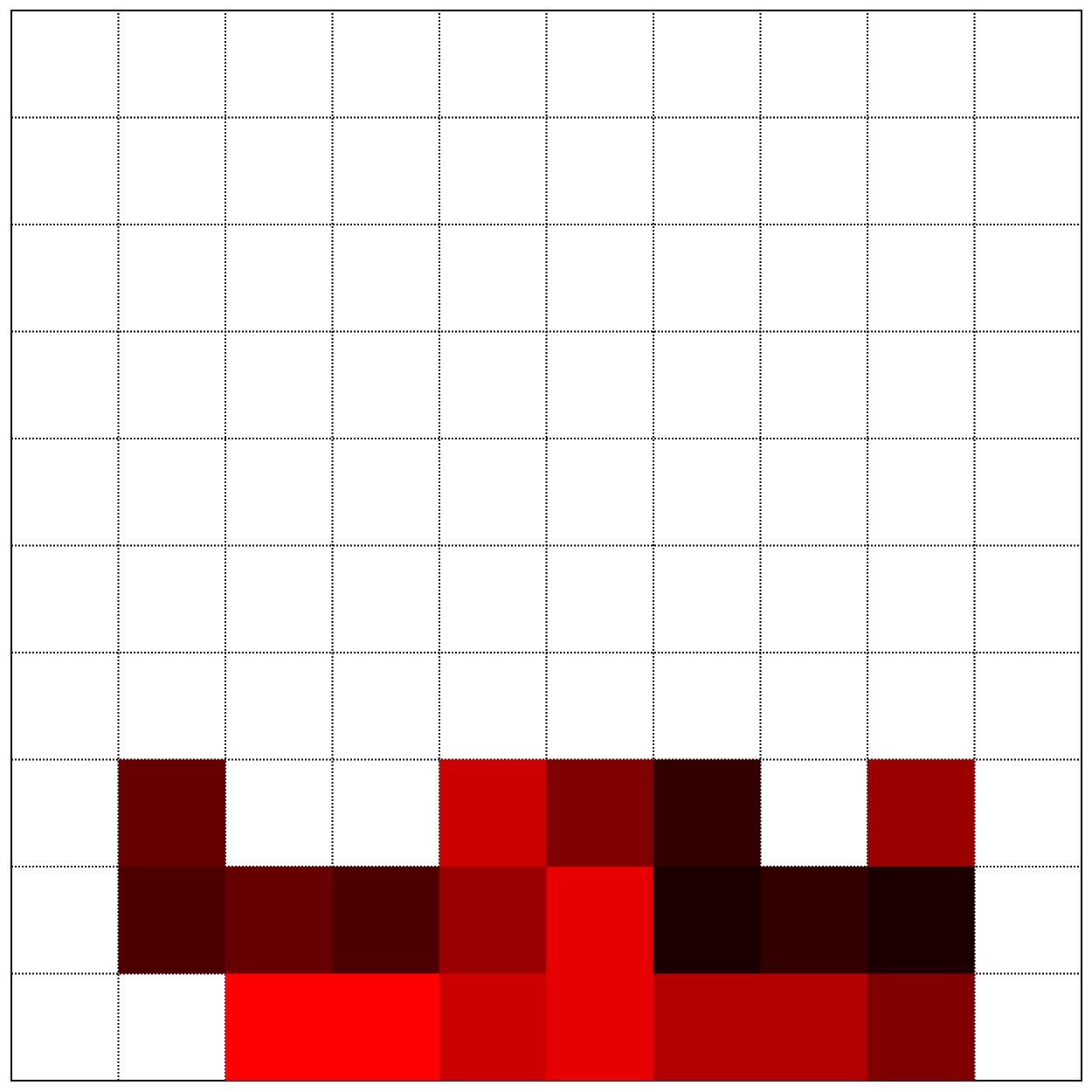}\label{fig9a}} 
	\hspace{3mm}
	\subfigure [Simulation] 
	{\includegraphics[scale=0.05]{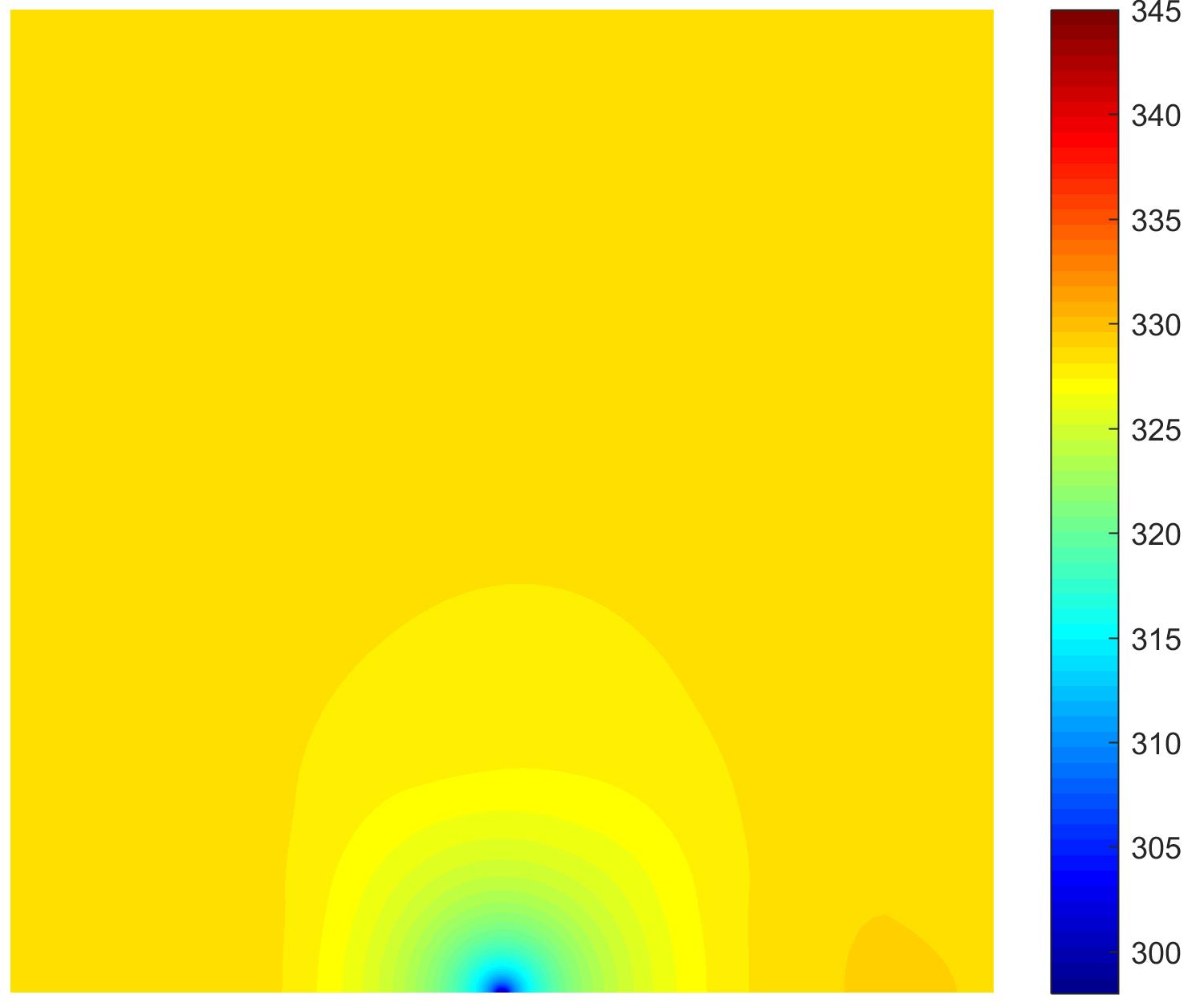}\label{fig9b}}
	\hspace{3mm}
	\subfigure [Layout]
	{\includegraphics[scale=0.05]{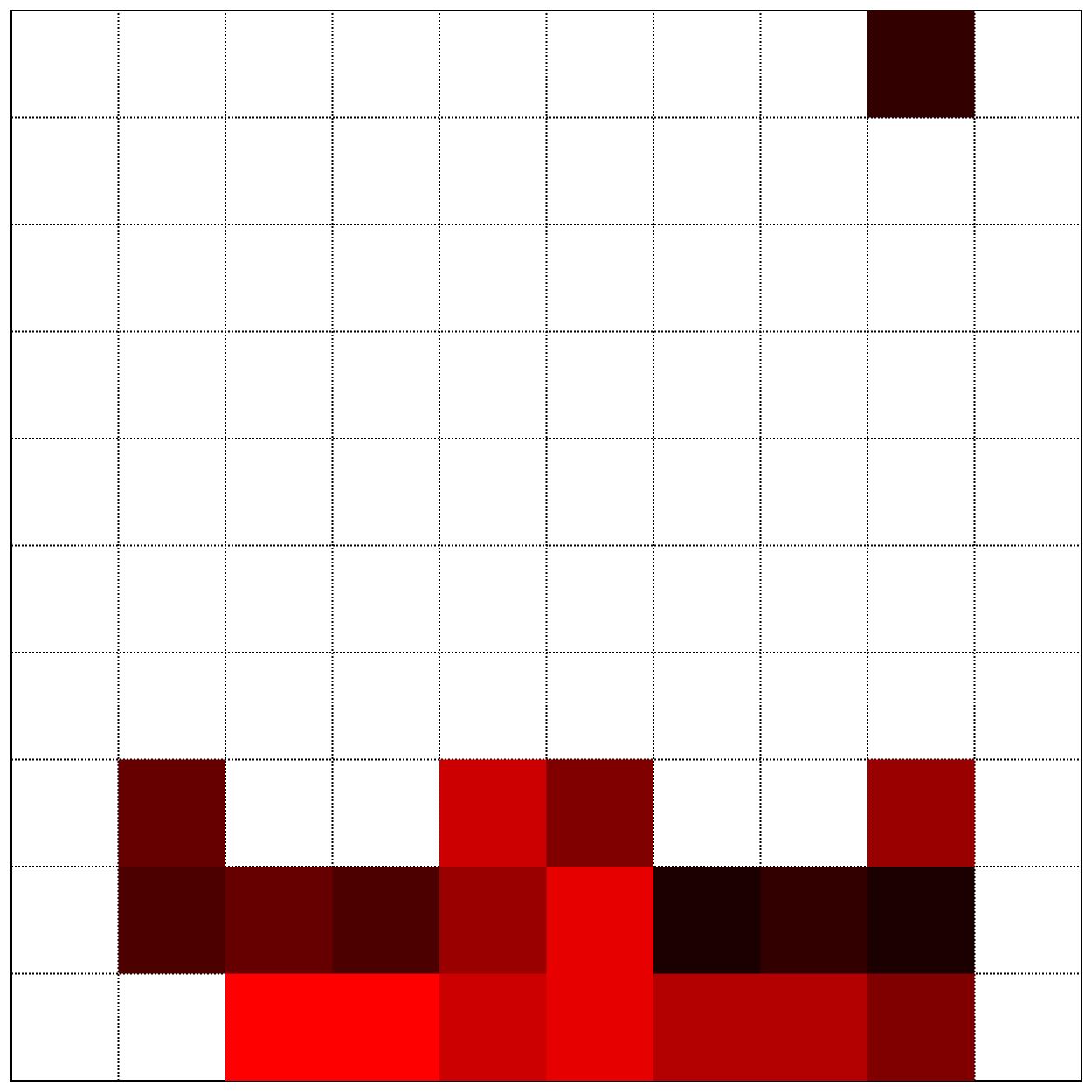}\label{fig9b}}
	\hspace{3mm}
	\subfigure [Simulation]
	{\includegraphics[scale=0.05]{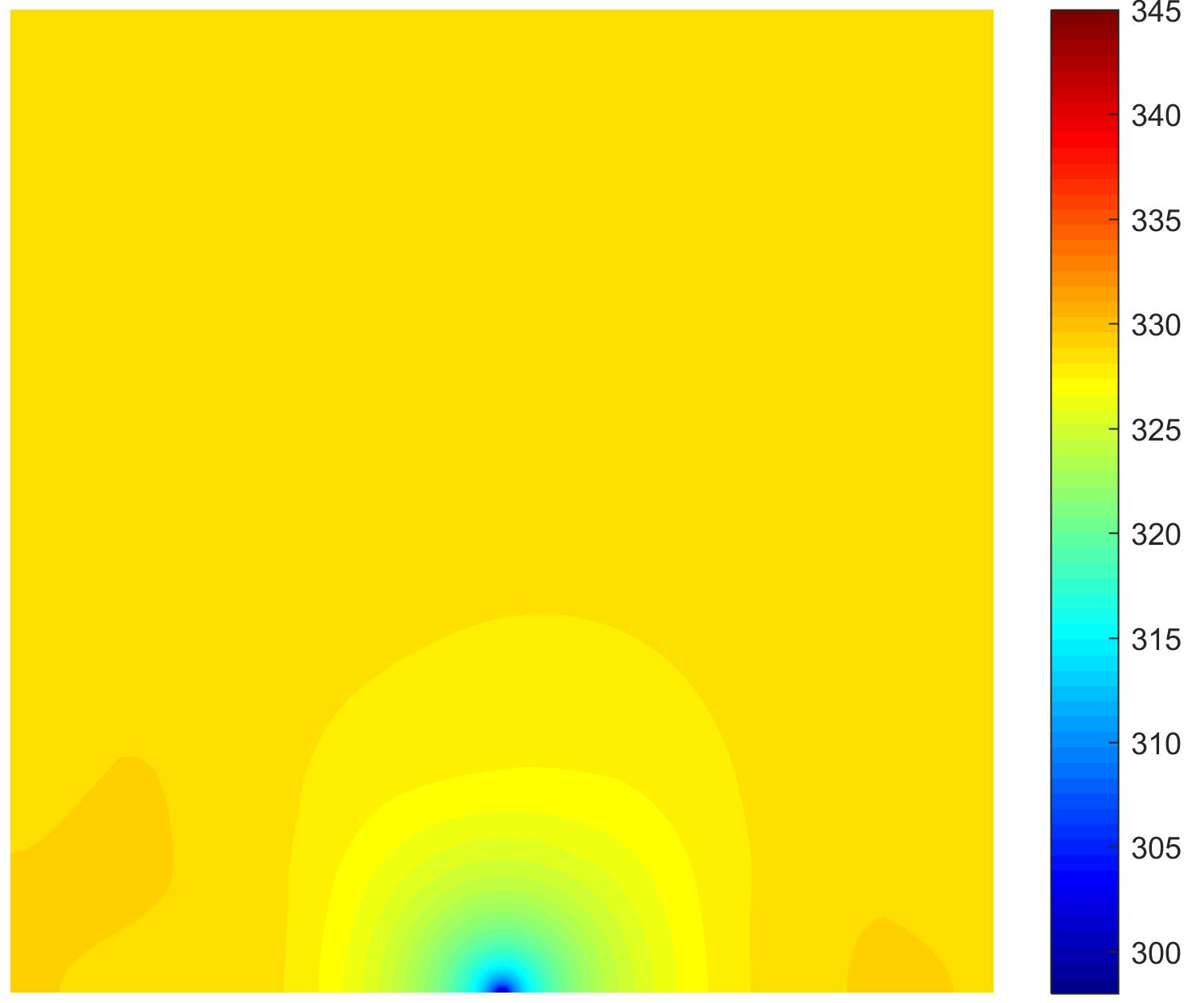}\label{fig9b}}
	
	\subfigure [Layout]
	{\includegraphics[scale=0.05]{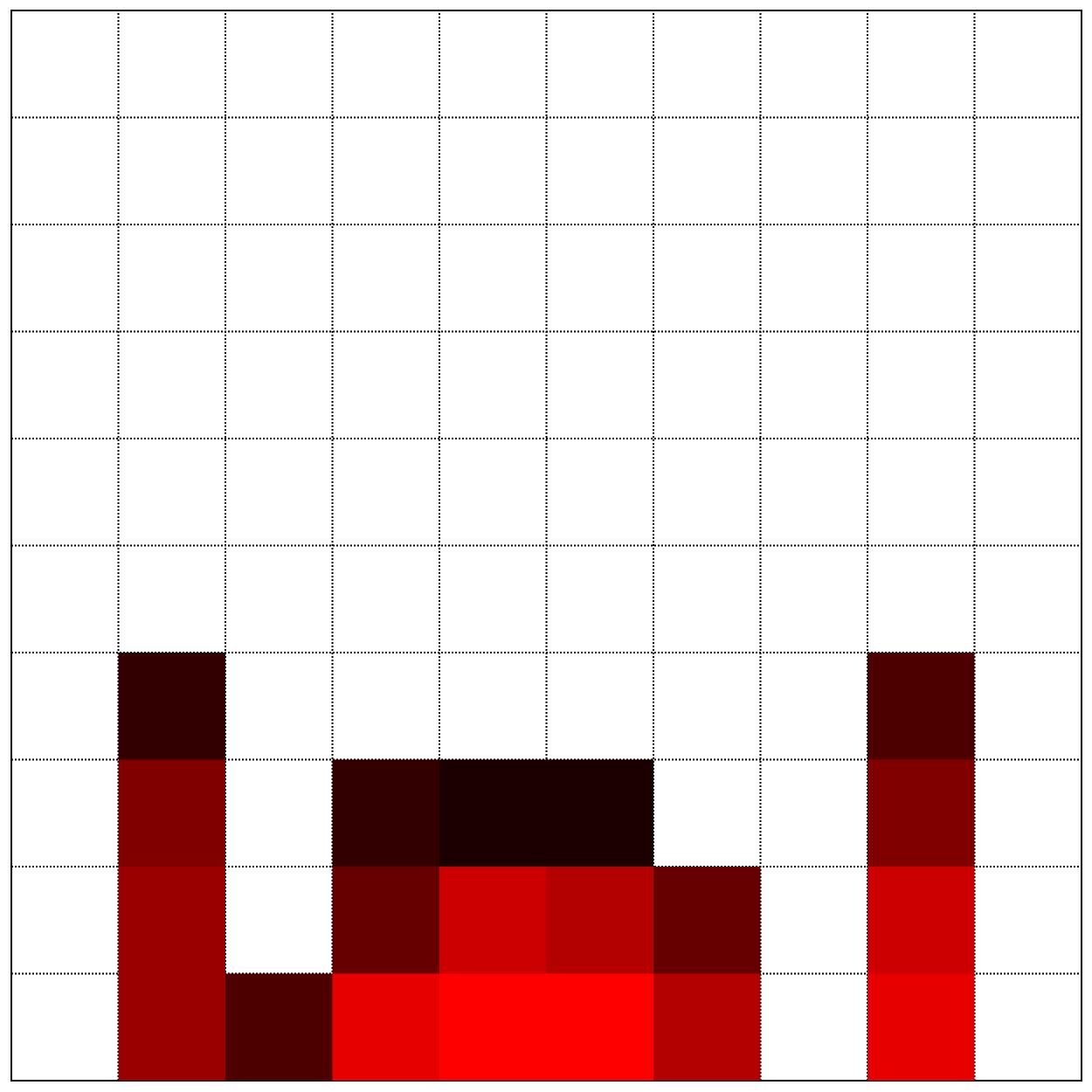}\label{fig9a}} 
	\hspace{3mm}
	\subfigure [Simulation] 
	{\includegraphics[scale=0.05]{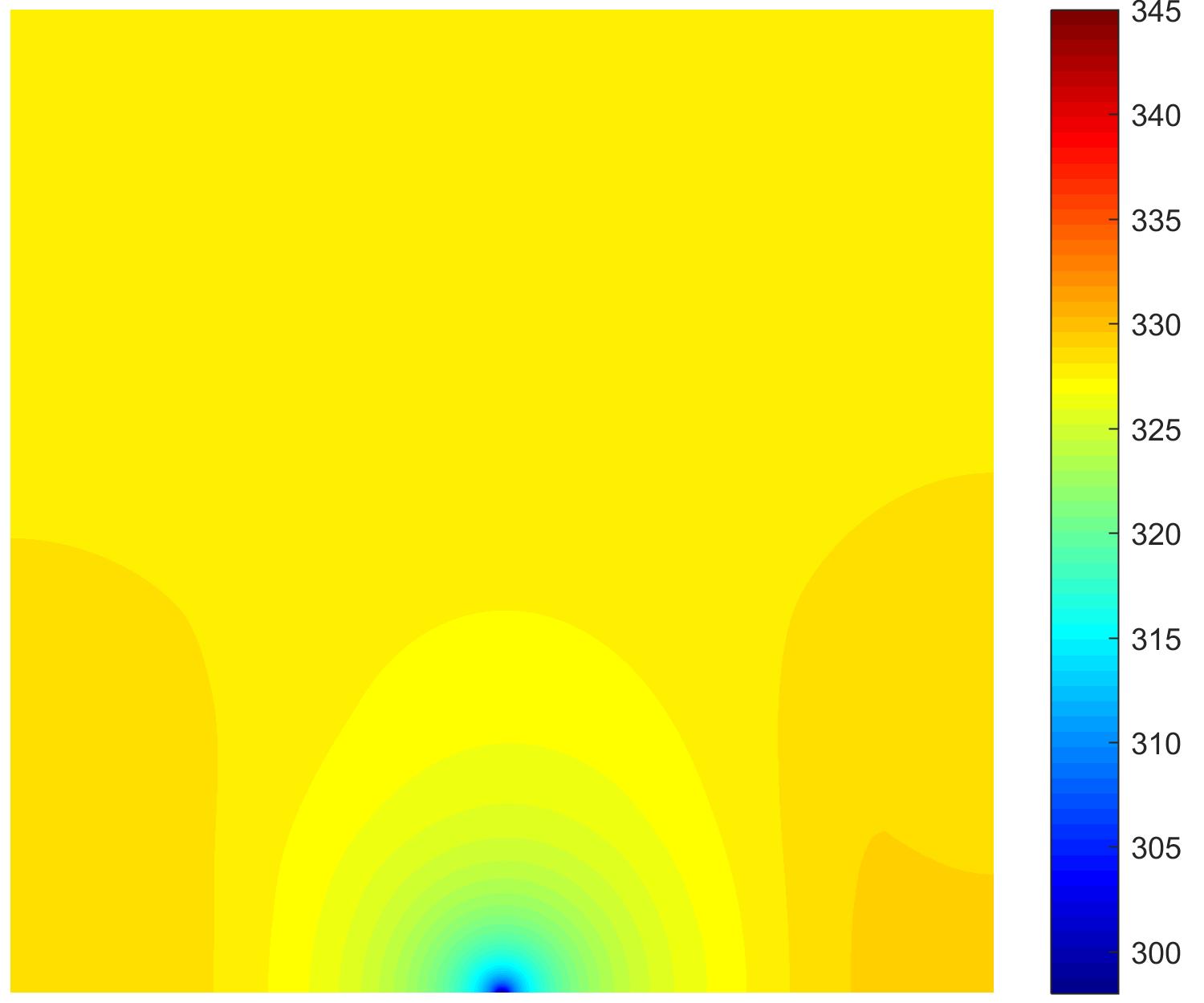}\label{fig9b}}
	\hspace{3mm}
	\subfigure [Layout]
	{\includegraphics[scale=0.05]{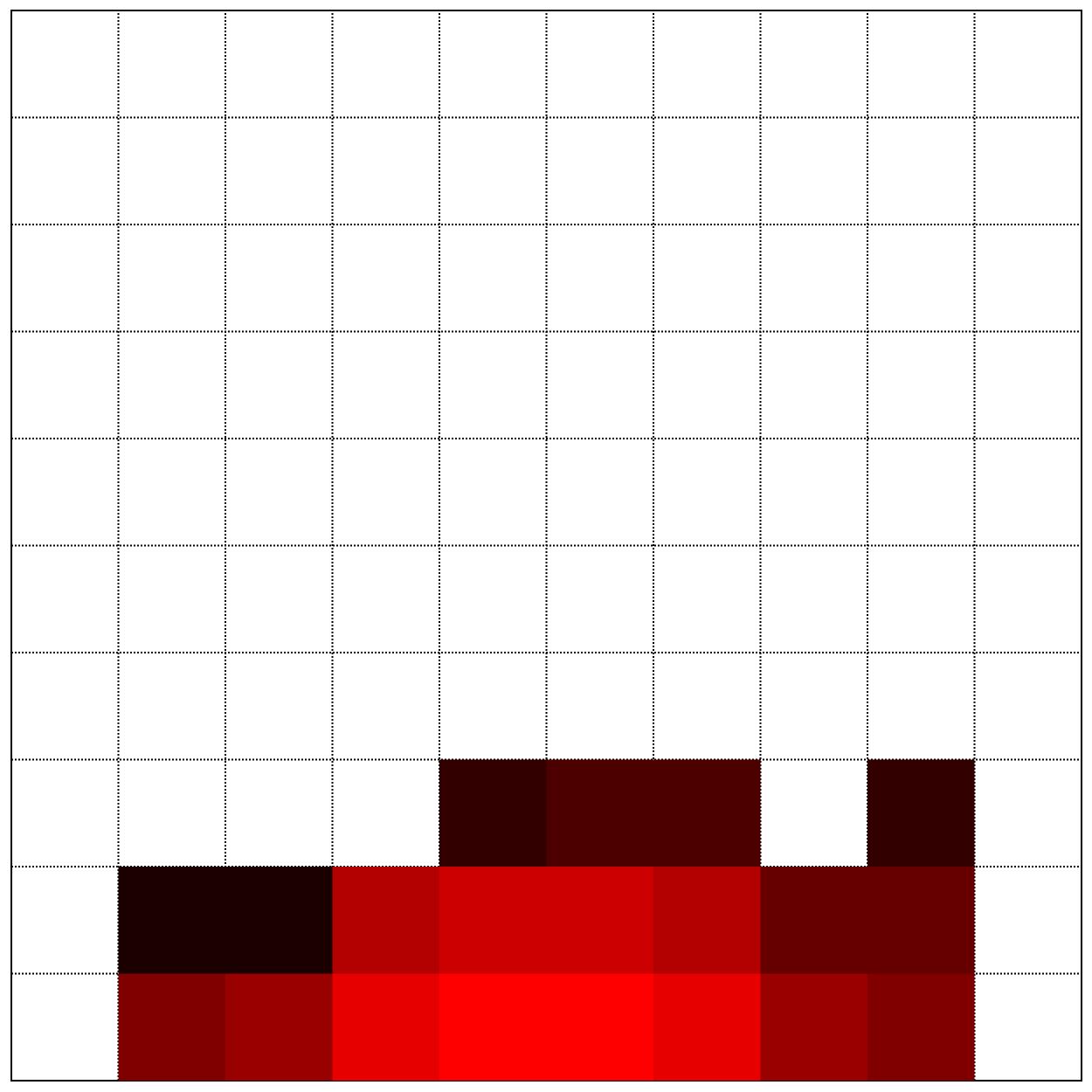}\label{fig9b}}
	\hspace{3mm}
	\subfigure [Simulation]
	{\includegraphics[scale=0.05]{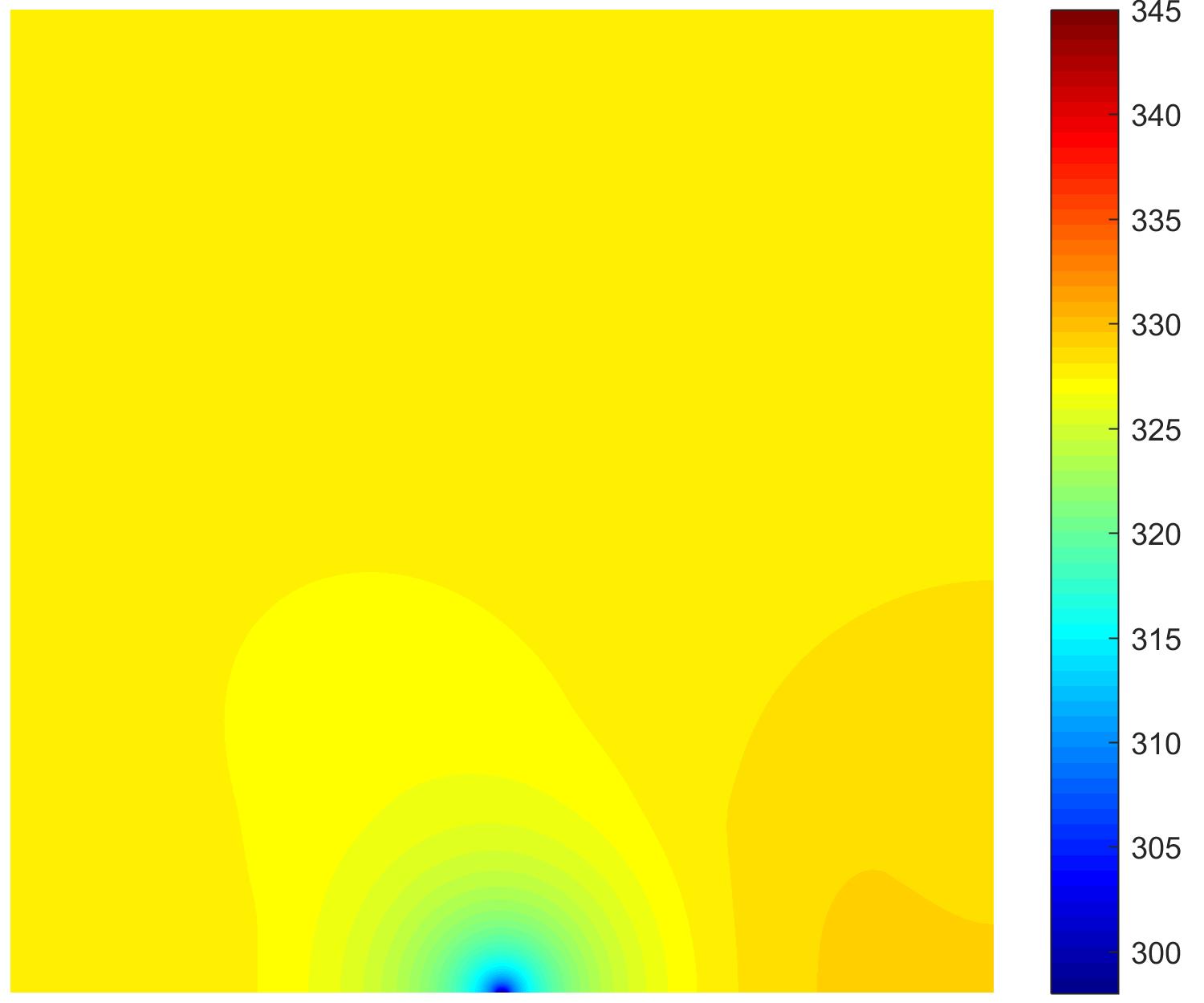}\label{fig9b}}
	\caption{An illustration of  four heat source layout schemes obtained by our method and their corresponding simulation in  case 2. Left top is MHSLO-NAS(5), Right top is MHSLO-NAS(6), Left bottom is MHSLO-NAS(7), Right bottom is MHSLO-NAS(8) }\label{fig17}
\end{figure*}

%
%
%
%

\subsubsection{MHSLO based on  the searched model}

Similar to case 1, after obtaining a deep learning surrogate model,  we use our designed MNSLO to identify the optimzal heat source layout scheme in case 2. To verify the effectiveness and improvement of the proposed MNSLO, we test the performance of multimodal optimization. We still implement it to solve case 2 to make a comparision. 

\begin{figure}[h]
	\centering
	\includegraphics[scale=0.06]{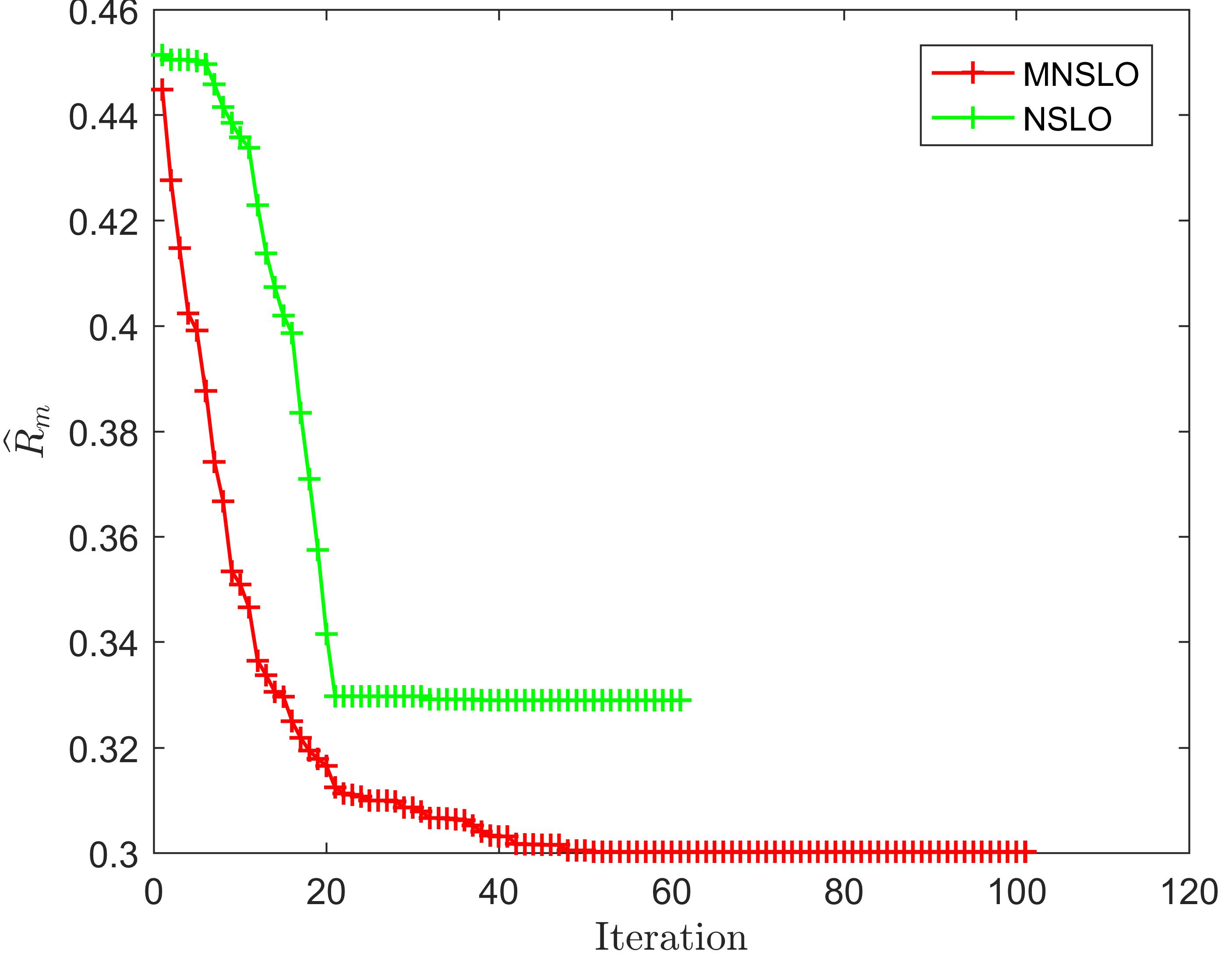}
	\caption{The iteration history of the NSLO algorithm and MNSLO algorithm for solving the case 2 based on Mixpath$\_$FPN surrogate.}
	\label{Iteration_NS2}
\end{figure}

To farily compare the effect of NSLO and MNSLO, the deep learning surrogate are both selected as the same Mixpath$\_$FPN model in two experiments. To compare the global optimziation ability, we set the number of groups $c$ in MNSLO as 1, the convergence curves of two algorithms are presented in Figure \ref{Iteration_NS2}. As we can see, MNSLO finds farther better solution than NSLO. NSLO is trapped in local optimum after 20 iterations. The maximum temperature of the simulation is only 333.51K, which is listed in Table \ref{Tab4}. The max temperature of founded heat source layout by MNSLO is reduced to 328.89K. From the obtained approximate optimal layout presented in Figure \ref{fig17}, it is reasonable that the higher intensity heat source should be located near the tiny heat sink. It means that our proposed algorithm could be also effective in more complex heat source layout task.

\begin{figure}[h]
	\centering
	\includegraphics[scale=0.9]{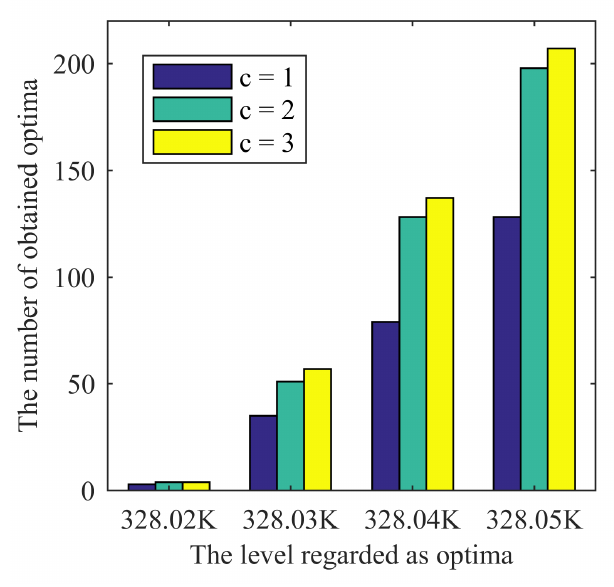}
	\caption{The number of the obtained layout design schemes in case 2 by MNSLO.}
	\label{mnslo1}
\end{figure}
Apart from obtaining the better solution compared with previous work, we further test the multimodal optimization effect of our proposed method. We set a threshld value of the maximum temperature as the level regarded as the optimal solutions. Then we evaluate the performance from the number of optimal solutions obtained by us. We set the threshold to 328.02K, 328.03K, 328.04K and 328.05K respectively, which are all lower than the result 333.51K implemented by Chen et al.\cite{Chen2020}. We also set the number of groups to 1, 2 and 3 to make a simple comparison. The result is presented in Figure \ref{mnslo1}. As we can see, when the threshold is set to 328.03K, we still could seek around fifty candidate solutions. We also find that when the number of groups is set to be larger, we could find more candidate optimal solutions. We list four searched layout schemes in Figure \ref{fig17} and Table \ref{Tab4}. As we can see, the maximum temperatures of these four layout schemes are very close. However, the difference between the layouts is relatively large. Thus our method could help to provide more design diversities for the designers.

\section{Conclusion}\label{sec8}
In this paper, the deep learning surrogate assisted HSLO method is further studied. Focusing on the two critical parts including the design of deep learning surrogate and the design of layout optimization algorithm, we propose a novel  framework of multimodal heat source layout optimization based on multi-objective neural architeture search.

From the aspect of constructing the deep learning surrogate, unlike the previous work to mannually design the neural network with rich debugging experience, we develop the neural architecture search method to automatically search for the optimal model architecture under the framework of FPN. Compared with the existing hand-crafted models, the searched model by us yields the state-of-art performance. With the similar arruracy, NAS finds models with 80$\%$ fewer parameters, 64$\%$ fewer FLOPs and 36$\%$ faster inference time than the original FPN model. 

From the aspect of optimization algorithm based on the deep learning surrogate, compared with previous work of only obtaining a local optimum in heat source layout optimization problem, we further design a multimodal neighborhood search based layout optimization algorithm to achieve multiple optimal solutions simultaneously. We utilize two cases to demonstrate and verify the perfoamance of our optimization algorithm. We achieve the state-of-art optimal layout schemes on both of two cases compared with other algorithms.In the first case heat source layout optimization problem with the same intensity, the maximum temperature of optimal layout is reduced from 327.02k to 326.74k. In addtion, our algorithm could provide almost one hundred similar layout, all of which are better than the result reported in the literture. In the second case with different intensity, our algorithm could find the layout scheme, the maximum temperature of which could reach 328.89k, farther lower than 333.51k than previous NSLO.

\backmatter

%
%
%

\bmhead{Acknowledgments}

This work was supported by the Postgraduate Scientific Research Innovation Project of Hunan Province (No.CX20200006) and the National Natural Science Foundation of China (Nos.11725211 and 52005505).

\bibliography{sn-bibliography}


\end{document}